\pdfoutput=1
\documentclass[11pt]{article}

\usepackage{emnlp2021}

\usepackage{times}
\usepackage{latexsym}

\usepackage[T1]{fontenc}
\usepackage[utf8]{inputenc}
\usepackage{enumitem}
\usepackage{microtype}

\usepackage{microtype}
\usepackage{url}
\usepackage{enumitem}
\usepackage{booktabs}
\usepackage{multirow}
\usepackage{array}
\usepackage{graphicx}
\usepackage{subcaption}
\usepackage{amsmath}
\usepackage{amsfonts}
\usepackage{amsthm,amssymb,amsopn,bm}
\usepackage{color,soul}
\usepackage{verbatim}
\usepackage{tabulary}
\usepackage{footnote}
\usepackage{blindtext}
\usepackage{algorithm}
\usepackage[noend]{algpseudocode}
\usepackage{stmaryrd}
\usepackage{makecell}
\usepackage{xcolor,colortbl}
\definecolor{green}{rgb}{0.1,0.1,0.1}
\definecolor{gitgreen}{HTML}{006400}

\makesavenoteenv{tabular}
\makesavenoteenv{algorithm}
\usepackage{tabularx,booktabs}
\newcolumntype{Y}{>{\centering\arraybackslash}X}
\newcolumntype{P}[1]{>{\centering\arraybackslash}p{#1}}

\usepackage{xcolor, soul}
\definecolor{lightyellow}{rgb}{1,1,0.7}
\sethlcolor{lightyellow}
\usepackage{xspace}

\newcommand\bolden[1]{{\boldmath\bfseries#1}}

\newcommand{\base}{\textsc{IndepPR}\xspace} % TODO
\newcommand{\sys}{\textsc{JPR}\xspace} % TODO
\newcommand{\decode}{\textsc{TreeDecode}\xspace} % TODO
\newcommand{\naive}{\textsc{SeqDecode}\xspace}
\newcommand{\dpr}{\textsc{DPR}$^+$}
\newcommand{\norerank}{\dpr\ only}
\newcommand{\nogueira}{\dpr+\citet{nogueira-etal-2020-document}}
\newcommand{\baseline}{\base}
\newcommand{\ours}{\sys}

\newcommand{\sota}{state-of-the-art}
\newcommand{\dev}{development}

\newcommand{\topk}{top $k$}
\newcommand{\rec}{\textsc{Recall @ $k$}}
\newcommand{\mrec}{\textsc{MRecall @ $k$}}

\newcommand{\nq}{\textsc{NQ}}
\newcommand{\ambigqa}{\textsc{AmbigQA}}
\newcommand{\webqsp}{\textsc{WebQSP}}
\newcommand{\trec}{\textsc{TREC}}

\newcommand{\eat}[1]{}

\newif\ifcomments
\commentsfalse

\providecommand{\sm}[1]{
    \ifcomments{\protect\color{purple!50!orange}{[Sewon: #1]}}\else
    \fi
}
\providecommand{\kl}[1]{
    \ifcomments{\protect\color{magenta}{[Kenton: #1]}}
    \else
    \fi
}
\providecommand{\mc}[1]{
    \ifcomments{\protect\color{violet}{[Mingwei: #1]}}\else
    \fi
}

\providecommand{\kt}[1]{
    \ifcomments{\protect\color{blue}{[Kristina: #1]}}\else
    \fi
}
\providecommand{\hanna}[1]{
    \ifcomments{\protect\color{purple}{[Hanna: #1]}}\else
    \fi
}
\providecommand{\commentout}[1]{}

\newcommand{\affilsup}[1]{\rlap{\textsuperscript{\normalfont#1}}}

\title{Joint Passage Ranking for Diverse Multi-Answer Retrieval}

\author{
    Sewon Min\affilsup{1},\thanks{~~Work done while interning at Google.}
    ~~Kenton Lee\affilsup{2},
    ~Ming-Wei Chang\affilsup{2},
    ~Kristina Toutanova\affilsup{2},
    ~Hannaneh Hajishirzi\affilsup{1} \\
     $^1$University of Washington \qquad
     $^2$Google Research \\
  \texttt{\{sewon,hannaneh\}@cs.washington.edu} \\
  \texttt{\{kentonl,mingweichang,kristout\}@google.com} \\
  }
\date{}

\begin{document}
\maketitle
\begin{abstract}
\sm{8p + Conclusion section currently}

\commentout{
\mc{I edited the abstract to be more align with current draft. please take a look.}\sm{

Updates:

- Figure 1

- 2nd paragraph of Section 1: Added two sentences contrasting with single-answer retrieval

- Section 2.3 (explicit mention that we challenge Izcard and Grave, footnote regarding the hardware constraint)

- Rename NaiveDecode to SeqDecode, SetDecode to TreeDecode

- Beginning of Section 3.1

- End of Section 3.3: added extreme cases when effective depth is 1 or $k$.

- Beginning of Section 4 (overview of Exp.)

- Section 4.2 (IndepPR)

- Beginning of Section 6: Big O notation for the memory added

}}

We study {\it multi-answer retrieval}, an under-explored problem that requires retrieving passages 
to cover multiple distinct answers for a given question.
%to have maximal coverage of a ground-truth set of multiple distinct answers. 
%Compared to single-answer passage retrieval,  multi-answer retrieval models need to score the retrieved passages jointly, as the models should not repeatedly retrieve passages containing the same answer and miss a different valid answer.
This task requires joint modeling of retrieved passages, as models should not repeatedly retrieve passages containing the same answer at the cost of missing a different valid answer.
% In this task, models must retrieve top relevant passages  while keeping them diverse.
%The task is important both as an end application where passages are presented to users and for downstream question answering where passages are consumed by an answer generation model.
%Prior retrieval systems such as DPR~\cite{karpukhin2020dense} is designed for single-answer retrieval and cannot reason about the set of passages jointly.
%Prior work focusing on single-answer retrieval is limited as it cannot reason about the set of passages jointly.
%\sm{Updated}
In this paper, we introduce \sys, the first joint passage retrieval model for multi-answer retrieval. % focusing on reranking.
%\kt{It is ambiguous what ``the first'' qualifies so might want to take out the reranking part as a separate sentence or just remove it because the next sentence mentions reranking.  }
\sys makes use of an autoregressive reranker that selects a sequence of passages, each conditioned on previously selected passages.
\sys\
%uses a training method that addresses the lack of supervision for the ordering of relevant passages,
is trained to select passages that cover new answers at each timestep and uses
a tree-decoding algorithm to enable flexibility in the degree of diversity.\kl{Did another edit to be more specific.}
%\sys\ comes with a training method that deals with the lack of supervision for the ordering of relevant passages, and a novel inference algorithm that enables flexibility in a degree of diversity.
%To address lack of supervision of ordering and be robust to early prediction errors, \sys\ comes with novel training and decoding algorithms.
%equipped with novel training and decoding algorithms \hanna{what is novel about them?}.
%dynamic oracle training, and a novel decoding algorithm.
Compared to prior approaches, \sys\ achieves significantly better answer coverage on three multi-answer datasets.
%with diverse retrieval results on three datasets\kt{Would it be better to say ... improved answer coverage and diversity of retrieved results ...}.
When combined with downstream question answering, the improved retrieval enables larger answer generation models since they need to consider fewer passages, establishing a new \sota.

\end{abstract}

\section{Introduction}\label{sec:intro}
\begin{figure}[t]
\centering
\resizebox{\columnwidth}{!}{\includegraphics[width=\textwidth]{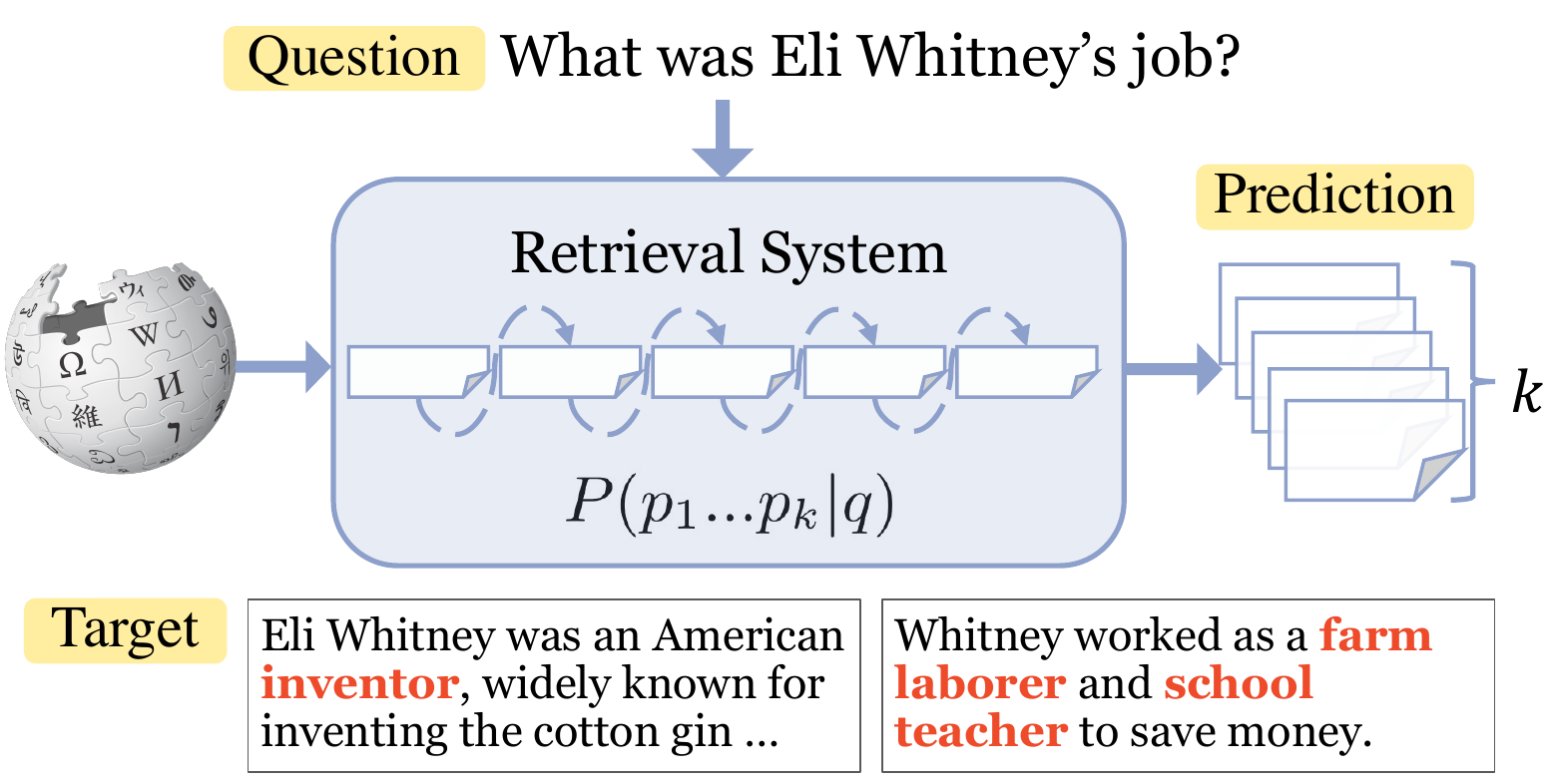}}
\caption{
    The problem of multi-answer retrieval.
    A retrieval system must retrieve a set of $k$ passages ($k=5$ in the figure) which has maximal coverage of diverse answers to the input question: {\em inventor}, {\em farm laborer} and {\em school teacher} in this example.
    This requires modeling the joint probability of the passages in the output set: $P(p_1...p_k|q)$.
    Our proposed model \sys\ achieves this by employing an autoregressive model.
}
\label{fig:intro}\end{figure}

Passage retrieval is the problem of retrieving a set of passages relevant to a natural language question from a large text corpus.
%The problem of passage retrieval requires retrieving a set of passages relevant to a natural language question given  a large text corpus.
Most prior work focuses on single-answer retrieval, which scores passages independently from each other according to their relevance to the given question, assuming there is a single answer~\citep{voorhees1999trec,chen2017reading,lee2019latent}.
However, questions posed by humans are often open-ended and ambiguous, leading to multiple valid answers~\citep{min2020ambigqa}.
For example, for the question in Figure~\ref{fig:intro}, ``What was Eli Whitney's job?'', an ideal retrieval system should provide passages covering all professions of Eli Whitney.
This introduces the problem of {\it multi-answer retrieval}---retrieval of multiple passages with maximal coverage of all distinct answers---which is a challenging yet understudied problem.

Multi-answer retrieval poses two challenges
that are not well represented in single-answer retrieval. First, the task requires scoring passages jointly to optimize for retrieving multiple relevant-yet-complementary passages. %, of which current models are not capable.
%making the problem difficult for current retrieval models.
%Second, the set retrieval property means that there is no gold ordering between the correct passages, and this property can be challenging to language-model-based models which assume the ordering is given during training.
Second, the model needs to balance %know when to explore, balancing
between two different goals: retrieving passages dissimilar to each other to increase the recall, and keeping passages relevant to the question.
%two very different strategies: one is to retrieve multiple similar passages to ensure at least one of them is relevant to the question, while the other one is to retrieve dissimilar passages to increase the recall. 

In this work, we introduce Joint Passage Retrieval (\sys), a new model that addresses these challenges.
To jointly score passages, \sys\ employs an encoder-decoder reranker and autoregressively generates passage references by modeling the probability of each passage as a function of previously retrieved passages.
Since there is no ground truth ordering of %relevant
passages, we employ a new training method that dynamically forms supervision to drive the model to prefer passages with answers not already covered by previously selected passages. Furthermore, we introduce a new tree-decoding algorithm to allow flexibility in the degree of diversity.
\kl{Reusing phrasing in abstract. I think we want to avoid saying that diversity and relevance is a tradeoff because (when the example requires it) more diversity should increase relevance.}
\mc{can we find another word for diversity?}

\commentout{

\hanna{still not happy with this paragraph}\sm{Updated}
%In this work, we study the multi-answer retrieval problem and introduce a new model for Joint Passage Retrieval (\sys) that captures dependencies between passages in the retrieved set by training and decoding  an autoregressive model. In particular, \sys\ learns to score passages using an encoder-decoder reranker that autoregressively generates a sequence of passage references given a set of candidate passages. We employ a training strategy that encourages multiple ways that multi-answer supervision can be reached with the reranker. Moreover,  we introduce a new tree-decoding algorithm for the autoregressive model, enabling flexibility in handling the trade-off between diversity of the top-$k$ passages and their relevance to the question. 
In this work, we introduce Joint Passage Retrieval (\sys), a new model that captures dependencies between passages in the retrieved set.
%\sys\ employs an encoder-decoder reranker that autoregressively generates a sequence of passage references given a set of candidate passages.
\sys\ employs an encoder-decoder reranker and autoregressively generates passage references by modeling the probability of each passage as a function of previously retrieved passages.
Since there is no groundtruth ordering of relevant passages, we employ a new training method that dynamically forms supervision to drive the model to prefer passages with answers not already covered by previously selected passages.
%Furthermore, we introduce a new tree-decoding algorithm that enables the model to be robust to early prediction errors.
Furthermore, we introduce a new tree-decoding algorithm that enables more flexible decoding based on a degree of diversity required for each question.
\kt{Do we have evidence the decoding helps with robustness to early errors? Previously we had said it helps balance relevance and diversity.} \sm{Haven't done direct analysis; the only evidence is that the performance is better... but IMO, I found the balance between relevance and diversity less convincing. Taking the top $k$ from the same step does not always improve the relevance. It helps relevance only when the top 1 prediction was incorrect so the top $k$ gives a chance to recover the error. So I think the more precise intuition is the balance between diversity and `recovering early errors'. And IMO having better performance by reducing the total steps does mean predictions from earlier steps got better, so I thought the result alone can be evidence.
} \kl{I think early error recovery is an over-general explanation of the behavior that you are seeing. Earlier steps are almost necessarily more accurate because of how the task is set up. We've trained the model to predict a passage at the Nth timestep that covers an answer distinct from N-1 other answers. If such an answer does not exist (which is more likely as N gets larger), then the model has no chance of being correct, regardless of the quality of the model. I would vote for reverting to the discussion about diversity. Perhaps a better way to discuss it is not that there is a direct trade-off between relevance and diversity since as you mentioned top-k from the same step is not necessarily more accurate. We can argue from a design point of view. The \emph{semantics} of the next step is that it's a passage that covers a different answer, but we want to back off other passages when there are no more new answers (i.e. there is no more useful diversity left.}

}
%Training and decoding of \sys\ is non-trivial due to lack of supervision of ordering of the passages and sensitiveness to early prediction errors.
%To this end, we employ a training strategy that encourages multiple ways that multi-answer supervision can be reached, and introduce a new tree-decoding algorithm that enables the model to be robust to early prediction errors.

%In this work, we study the multi-answer retrieval problem and introduce a new model for Joint Passage Retrieval (\sys) that captures dependencies between passages in the retrieved set leveraging an autoregressive model. In particular,     
%\sys\ uses a sequence-to-sequence reranker, receiving a set of candidate passages and autoregressively generating a sequence of passage references.
%jThe autoregressive behavior of the model can capture the dependencies between passages in the output set.
%We employ a training method that encourages multiple ways to reach the multi-answer supervision, and introduce a new tree-decoding algorithm for the autoregressive model, enabling flexibility in handling the trade-off between diversity of the top-$k$ passages and their relevance to the question. \hanna{are we evaluating diversity?}

In a set of experiments on three multi-answer datasets---\webqsp~\citep{yih2016value}, \ambigqa~\citep{min2019compositional} and \trec~\citep{baudivs2015modeling}, \sys\ achieves significantly improved recall over both a dense retrieval baseline~\citep{guu2020realm,karpukhin2020dense} and a \sota\ reranker that independently scores each passage~\citep{nogueira-etal-2020-document}.
Improvements are particularly significant on questions with more than one answer, outperforming dense retrieval by up to 12\% absolute and an independent reranker by up to 6\% absolute.

We also evaluate the impact of \sys\ in downstream question answering, where an answer generation model takes the retrieved passages as input and generates short answers. Improved reranking leads to improved answer accuracy because we can supply fewer, higher-quality passages to a larger answer generation model that fits on the same hardware.
This practice leads to a new \sota\
on three multi-answer QA datasets and \nq~\citep{kwiatkowski2019natural}. To summarize, our contributions are as follows:

%\sm{To remove bullet points, we'll need concluding sentences that put the story back to multi-answer from NQ;} \hanna{let's not remove bullet points yet; I like them}

\begin{enumerate}[topsep=1pt,itemsep=-.8mm] %\setlength\itemsep{-.2em}
    \item We study multi-answer retrieval, an underexplored problem that requires the \topk\  passages to maximally cover the set of distinct answers to a natural language question.
    \item We propose \sys, a joint passage retrieval model that integrates dependencies among selected passages, along with new training and decoding algorithms.
    \item On three multi-answer QA datasets, \sys\ significantly outperforms a range of baselines with independent scoring of passages, both in retrieval recall and answer accuracy.
\end{enumerate}

\section{Background}\label{sec:task}
\subsection{Review: Single-Answer Retrieval}
In a typical single-answer retrieval problem, a model is given a natural language question $q$ and retrieves $k$ passages $\{p_1...p_k\}$ from a large text corpus $\mathcal{C}$~\citep{voorhees1999trec,ramos2003using,robertson2009probabilistic,chen2017reading,lee2019latent,karpukhin2020dense,luan2020sparse}. The goal is to retrieve at least one passage that contains the answer to $q$. During training, question-answer pairs $(q,a)$ are given to the model.

\vspace{-.3em}
\paragraph{Evaluation}
{\em Intrinsic} evaluation directly evaluates the retrieved passages.
The most commonly used metric is \rec\ which considers  retrieval successful if the answer $a$ is included in $\{p_1...p_k\}$.
{\em Extrinsic} evaluation uses 
the retrieved passages as input to an answer generation model such as the model in~\citet{izacard2020leveraging} and evaluates final question answering performance.

%\paragraph{Dense retrieval}
%State-of-the-art retrieval models use a dual-encoder architecture with supervised training~\cite{yih2011learning,karpukhin2020dense}. Such models learn to score passages with dot products between question ($q$) and passage ($p$) vectors, $$ P(p|q) \propto f_q(q)^T f_p(p),$$ where $f_q$ and $f_p$ are question and passage encoders, respectively, often from pre-trained BERT~\citep{devlin2019bert}. These models are highly scalable and can be applied to efficiently retrieve the highest-scoring passages from large collections.

\begin{table}[t]
\setlength{\tabcolsep}{8pt}
    \centering \footnotesize
   \begin{tabularx}{\columnwidth}{l@{\extracolsep{\fill}}l@{\extracolsep{\fill}}l}
        \toprule
            Task & \makecell[c]{\textbf{Single-answer}\\\textbf{Retrieval}} & \makecell[c]{\textbf{Multi-answer}\\\textbf{Retrieval}} \\
            \midrule
           Train Data 
             & $(q,a)$ & $(q,\{a_1...a_n\})$ \\
            \cmidrule{1-3}
            Inference & $q \rightarrow \{p_1...p_k\}$ & $q \rightarrow \{p_1...p_k\}$ \\
            \cmidrule{1-3}
            Evaluation & \makecell[l]{\textsc{Recall}$($\\
            ~~~$a,\{p_1...p_k\})$} & \makecell[l]{\textsc{MRecall}$($\\
            ~~~$\{a_1...a_n\},$\\
            ~~~$\{p_1...p_k\})$}\\
            \cmidrule{1-3}
            \makecell[l]{Appropriate\\Model} & $P(p_i|q)$ & $P(p_1...p_k|q)$  \\
        \bottomrule
    \end{tabularx}
    \caption{ A comparison of single-answer and multi-answer retrieval tasks. Previous work has used independent ranking models $P(p_i|q)$ for multi-answer retrieval because the inference-time inputs and outputs are the same.
    %This is the wrong tool for the problem, since it does not model dependencies within the output set.
    We propose \sys\ as an instance of $P(p_1...p_k|q)$. % that is more appropriate for multi-answer retrieval.
    }\label{tab:compare-sar-mar}
\end{table}

\begin{figure*}[t]
\centering
\resizebox{1.8\columnwidth}{!}{\includegraphics[width=1.8\textwidth]{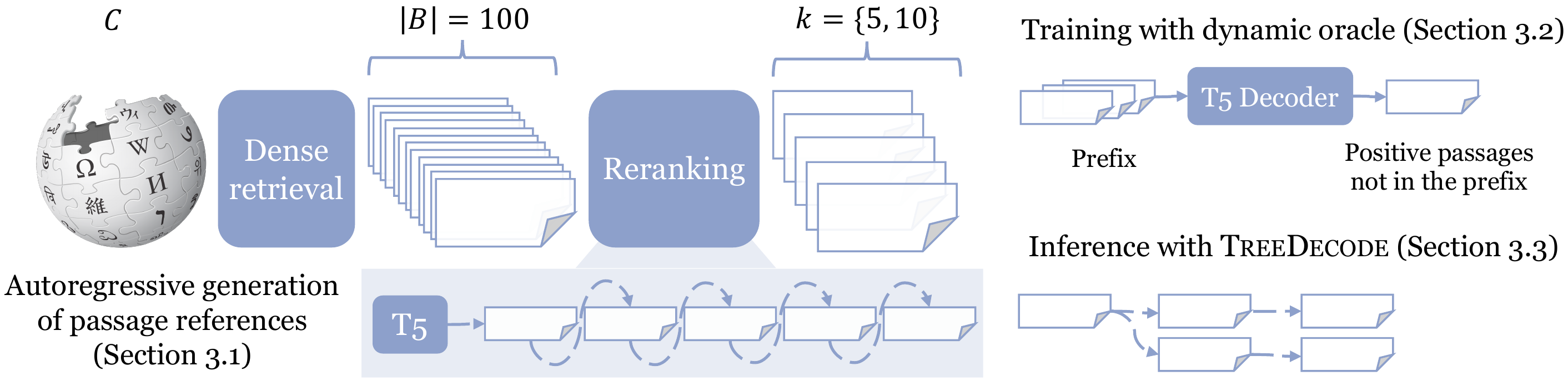}}
\caption{
    An overview of \sys. %, consisting of dense retrieval and reranking.
    We focus on reranking and propose: autoregressive generation of passage references (Section~\ref{subsection:autoregressive}), training with dynamic oracle (Section~\ref{subsection:training}), and inferece with \decode\ (Section~\ref{subsection:setdecode}).
}
\label{fig:overview}\end{figure*}

\vspace{-.3em}
\paragraph{Reranking} 
Much prior work~\citep{liu2011learning,asadi2013effectiveness,nogueira-etal-2020-document} found an effective strategy in using a  two-step approach of (1) retrieving a set of candidate passages $\mathcal{B}$ from the corpus $\mathcal{C}$ ($k < |\mathcal{B}| \ll |\mathcal{C}|$) %through lexical matching or a dense retrieval model
and (2) using another model to rerank the passages, obtaining a final top $k$. 
A reranker could be more expressive than %a dual encoder,
the first-stage model (e.g. by using cross-attention), as it needs to process much fewer candidates.
%e.g., having cross-attention between the question and the passage, as it deals with much fewer passages.
Most prior work in reranking, including the current \sota~\citep{nogueira-etal-2020-document}, scores each passage independently, modeling $P(p|q)$.
%models an independent score of each passage %$P(p|q)$.

\subsection{Multi-Answer Retrieval}\label{subsec:multi-answer-retrieval}
We now formally define the task of multi-answer retrieval. A model is given a natural language question $q$ and needs to find $k$ passages $\{p_1...p_k\}$ from $\mathcal{C}$ that contain {\em all distinct answers} to $q$.
Unlike in single-answer retrieval,  question-and-answer-set pairs $(q,\{a_1...a_n\})$ are given during training.

\vspace{-.3em}
\paragraph{Evaluation}
Similar to single-answer retrieval, the {\em intrinsic} evaluation directly evaluates a set of $k$ passages.
As the problem is underexplored,  metrics for it are less studied. We propose to use \mrec, a new metric which considers retrieval to be successful if all answers or at least $k$ answers in the answer set $\{a_1...a_n\}$ are recovered by $\{p_1...p_k\}$.
Intuitively, \textsc{MRecall} is an extension of \textsc{Recall} that considers the completeness of the retrieval; the model must retrieve all $n$ answers when $n \leq k$, or at least $k$ answers when $n > k$.\footnote{
    This is to handle the cases where $n$ is very large (e.g. over 100) and the model covers a reasonable number of answers given the limit of $k$ passages, therefore deserves credit.}
    
The {\em extrinsic} evaluation inputs
the retrieved passages into an answer generation module that is designed for
%generating
multiple answers,
and measures multi-answer accuracy using an appropriate metric such as the one in~\citet{min2020ambigqa}.
%the accuracy of them,
%We then evaluate the performance of the end-to-end QA performance using the metric proposed in each dataset such as \ambigqa~\cite{min2020ambigqa}.
%using the metric such as the one in~\citet{min2020ambigqa}.

\vspace{-.3em}
\paragraph{Comparing to single-answer retrieval}
We compare single-answer retrieval and multi-answer retrieval in Table~\ref{tab:compare-sar-mar}.
Prior work makes no distinctions between these two problems,
since they share the same interface during inference.
%However, single-answer retrieval and multi-answer retrieval aim to achieve different goals and should be modeled differently.
%More precisely,
%While independently scoring each passage ($P(p_i|q)$) may be sufficient for single-answer retrieval, {\em joint} passage scoring ($P(p_1...p_k|q)$ is more appropriate for multi-answer retrieval.
However, while independently scoring each passage ($P(p_i|q)$) may be sufficient for single-answer retrieval, multi-answer retrieval inherently requires {\em joint} passage scoring $P(p_1...p_k|q)$.
For example, %in the multi-answer setting,
models should not repeatedly retrieve the same answer at the cost of missing other valid answers,
which can only be done by a joint model.
%; only a joint model has the capacity to model the relevance between $q$ and $p_i$ and the diversity among $\{p_1...p_k\}$ at the same time.

\vspace{-.3em}
\paragraph{Choice of \bolden{$k$} for downstream QA}
Previous \sota\ models typically input
%a large number of passages ($k=100$)
a large number ($k \geq 100$) of passages
to the answer generation model. For instance, \citet{izacard2020leveraging} claim the importance of
using a larger value of $k$ to improve QA accuracy.
In this paper, we argue that with reranking, using a smaller value of $k$ (5 or 10) and instead employing a larger answer generation model is advantageous
given a fixed hardware budget.\footnote{
We care about a fixed type of hardware
%because in retrieval and open-domain QA research, the available hardware with limited memory
since it is the hardest constraint and usually a bottleneck for performance. We do not control for running time in this comparison. }
We show in Section 5 that, as retrieval performance improves, memory is better spent on larger answer generators rather than on more passages, ultimately leading to higher QA accuracy.

\section{\sys: Joint Passage Retrieval}\label{sec:model}\begin{figure*}[ht]
\centering
\resizebox{1.7\columnwidth}{!}{\includegraphics[width=1.9\textwidth]{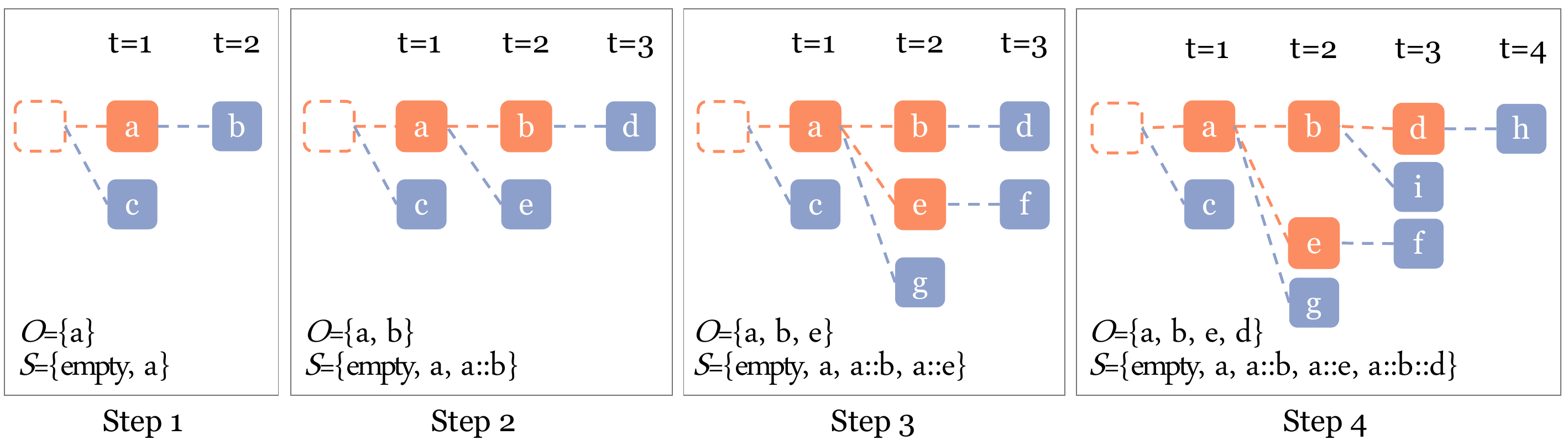}}
\caption{
    An illustration of \decode, where passages that are chosen and passages that are being considered are indicated in orange and blue, respectively.
    See Section~\ref{subsection:setdecode} and Algorithm~\ref{alg:decoding} for details. \hanna{it might be good if you show the output of seqdeocode or independent decoding}
}
\label{fig:decoding}\end{figure*}

\sm{This section is generally all updated.}

We propose \sys\ (Joint Passage Retrieval), which models the joint probability $P(p_1...p_k|q)$ for multi-answer retrieval.
JPR uses an approach consisting of first-stage retrieval followed by reranking: the first-stage retrieval obtains candidate passages $\mathcal{B}$ from the corpus $\mathcal{C}$, and a reranker processes $\mathcal{B}$ to output $\{p_1...p_k\} \subset \mathcal{B}$.
%\hanna{you cannot only refer to the appendix; need to mention something here, refer to the background section that you've introduced DPR; if anything important has changed over it and hence you've called it DPR+ then you need to mention. Or, simply you can say you extend DPR to our problem setup (refer to the appendix for the details.}
We refer to Section~\ref{subsec:dense-retrieval} for the first-stage retrieval, and focus on the reranking component of the model,
which allows 
(1) efficiently modeling the joint probability $P(p_1...p_k|q)$, and
(2) processing candidate passages with a more expressive model.

The overview of \sys\ is illustrated in Figure~\ref{fig:overview}.
The reranker of \sys\ leverages the encoder-decoder architecture for an autoregressive generation of passage references (Section~\ref{subsection:autoregressive}).
Unlike typical use cases of the encoder-decoder, (1) the ordering of passages to retrieve is not given as supervision, and (2) 
it is important to balance between exploring passages about new answers %to increase the recall,
and finding passages that may cover previously selected answers.
%and keeping all retrieved passages relevant to the question.
%it is not necessary to derive the output set from a single sequence which may cause the output to be sensitive to early prediction errors.\kl{Same issue regarding early prediction errors. I think this will raise red flags to the reviewers because it's not thoroughly explained.}
To this end, we introduce a new training method (Section~\ref{subsection:training}) and a tree-based decoding algorithm (Section~\ref{subsection:setdecode}).

\commentout{
We propose \sys\ (Joint Passage Retrieval), which models the joint probability of $P(p_1...p_k|q)$ for multi-answer retrieval.
Formally, a model is given a question $q$ and outputs a set of $k$ passages $\mathcal{O} = \{o_1...o_k\}$ from a large text corpus $\mathcal{C}$; $\mathcal{O}$ should have maximal coverage of the answer set to $q$.

We use an approach consisting of dense retrieval and reranking: dense retrieval retrieves candidate passages $\mathcal{B}$ from $\mathcal{C}$, and a reranker processes $\mathcal{B}$ to output $\mathcal{O} = \{o_1...o_k\} \subset \mathcal{B}$. 
Often $k < |\mathcal{B}| \ll |\mathcal{C}|$; we use $|\mathcal{B}|=100$ in our experiments.
In this work, we focus on the reranking component of the model, which allows (1) processing candidate passages with a more expressive model, and (2) efficiently modeling the joint probability of $P(p_1...p_k|q)$.
We refer to Appendix~\ref{app:dense-retrieval} for the details of dense retrieval, and describe reranking here.

The overview of \sys\ is illustrated in Figure~\ref{fig:overview}. Our reranking leverages the encoder-decoder architecture for an autoregressive generation of passage references (Section~\ref{subsection:autoregressive}). We employ dynamic oracle training to successfully train our T5-based architecture (Section~\ref{subsection:training}). Finally, we introduce an inference algorithm to decode a set of passage references (Section~\ref{subsection:setdecode}). These new training and inference methods are proposed in order to tackle the discrepancy between modeling a sequence and modeling a set.}

\commentout{\\

Leaving previous paragraphs as comment:

Our model benefits from the pretrained autoregressive decoding of T5..
We focus on reranking because recent studies have shown the importance of using large-scale pretrained models~\citep{devlin2019bert,raffel2020exploring}, and our proposed reranker can naturally benefit from the pretrained autoregressive decoding of T5. Applying T5 to multi-answer reranking is non-trivial \hanna{Why isn't it trivial? explain challenges and correspond your solutions to match those challenges.} , and \sys\ introduces three new concepts to make this possible (1) generating a sequence of references, \hanna{What isa  reference?} \hanna{generating a sequence of references to passages as their rank}(2) dynamic oracle training \hanna {how would this help?}, and (3) inference with \decode. \hanna{inference and decoding a set of passages with our \decode.}

\hanna{can't parse the sentence; you focus on reranking because previous models show benefit of T5?}
\hanna{I would rephrase this paragraph as follows: Our reranking uses a encoder-decoder architecture (e.g., T5) for reranking. Our model benefits from the pretrained autoregressive decoding of T5..}
We focus on reranking because recent studies have shown the importance of using large-scale pretrained models~\citep{devlin2019bert,raffel2020exploring}, and our proposed reranker can naturally benefit from the pretrained autoregressive decoding of T5. Applying T5 to multi-answer reranking is non-trivial \hanna{Why isn't it trivial? explain challenges and correspond your solutions to match those challenges.} , and \sys\ introduces three new concepts to make this possible (1) generating a sequence of references, \hanna{What isa  reference?} \hanna{generating a sequence of references to passages as their rank}(2) dynamic oracle training \hanna {how would this help?}, and (3) inference with \decode. \hanna{inference and decoding a set of passages with our \decode}
}

%\paragraph{Autoregressive Design.}
\subsection{Autoregressive generation of passage references}\label{subsection:autoregressive}

\sys\ makes use of the encoder-decoder architecture, where the encoder processes candidate passages
%following \citet{izacard2020leveraging}, \hanna{this izacard paper is breaking the flow? have you refered to this work before? if yes, just remove this sentence.}
and the decoder autoregressively generates a sequence of $k$ passage references ({\em indexes}). \hanna{are these passage references capture the rank of a passage?} \sm{No, we randomly shuffle them}
Intuitively, dependencies between passages can be modeled by the autoregressive architecture.%\footnote{This strategy of generating references can be considered an alternative to pointer networks~\citep{pointer_networks} that does not require any change to the existing pretrained encoder-decoder architecture.} %\hanna{maybe a footnote?} %\hanna{any autoregressive architecture would capture these dependencies? in that case, would Izacard and Grave system also able to do multipassage retrieval?}

%\hanna{big comment here: don't write forming input for the encoder in a procedural way, rather write it in a mathematical way, like The input to the encoder is formed by concatenating x with y with z...}
%First,
%\hanna{where is second?}
We extend the architecture from \citet{izacard2020leveraging}; we reuse the encoder but modify the decoder.
Each candidate passage $p_i$ is concatenated with the question $q$ and the number $i$ (namely {\em index}). It is fed into the encoder to be transformed to $\mathbf{p}_i \in \mathbb{R}^{L \times h}$, where 
$L$ is the length of the input text and $h$ is a hidden size.
Next, $\mathbf{p}_1...\mathbf{p}_{|\mathcal{B}|}$ are concatenated to form $\bar{\mathbf{p}} \in \mathbb{R}^{L|\mathcal{B}| \times h}$, and then fed into the decoder.
% \hanna{definitely remove T5; instead, write down something like this: mathbf{p} = enc{....}; and then an equation for decode.}
The decoder is trained to autoregressively output a sequence of indexes $i_1...i_k$, %($1 \leq i_1...i_k \leq |\mathcal{B}|$),
representing a sequence of passages $p_1...p_k$. As the generation at step $t$ ($1 \leq t \leq k$) is dependent on the generation at step $1\dots{t-1}$, it can naturally capture dependencies between selected passages.
As each index occupies one token, the length of the decoded sequence is $k$.

\subsection{Training with Dynamic Oracle}\label{subsection:training}
A standard way of training the encoder-decoder is teacher forcing which assumes a single correct sequence. However, in our task, a set of answers can be retrieved through many possible sequences of passages, and it is unknown which sequence is the best.
To this end, we dynamically form the supervision data which pushes the model to assign high probability to a {\em dynamic oracle}---any positive passage covering a correct answer that is not in the prefix, i.e., previously selected passages.
%In order to make the most of the multi-answer supervision, we sample passage prefixes during teacher forcing, and define a dynamic oracle that pushes the model to assign high probability to any positive passage covering a correct answer that is not in the prefix.

\hanna{Here, you need an opening statement saying you form positive/negative sequences this way; also it is good to connect this to the name that you are using: dynamic oracle; what is dynamic here.} \sm{Updated.}
We first precompute a set of positive passages $\mathcal{\Tilde{O}}$ and a prefix $\Tilde{p}_1...\Tilde{p}_k$.
A set of positive passages $\mathcal{\Tilde{O}}$ includes up to $k$ passages with maximal coverage of the distinct answers.\footnote{ $|\mathcal{\Tilde{O}}| < k$ if fewer than $k$ passages are sufficient to cover all distinct answers; $|\mathcal{\Tilde{O}}|=k$ otherwise.}
A prefix $\Tilde{p}_1...\Tilde{p}_k$ is a simulated prediction of the model, consisting of $\mathcal{\Tilde{O}}$ and $k-|\mathcal{\Tilde{O}}|$ sampled negatives.
%
%Given a set of positive passages $\mathcal{\Tilde{O}}$ and a prefix $\Tilde{p}_1...\Tilde{p}_k$, %\hanna{the prev. sentence is not clear until you read the next paragraph; I suggest to move the next paragraph before the objective.}  
%the dynamic oracle at each step $t$ ($1 \leq t \leq k$) is defined as
%$\mathcal{\Tilde{O}}-\{\Tilde{p}_1...\Tilde{p}_{t-1}\}$.
%, \sys\ takes $\Tilde{p}_1...\Tilde{p}_{t-1}$ as the decoder prefix and is trained to output the indexes of passages in $\mathcal{\Tilde{O}}-\{\Tilde{p}_1...\Tilde{p}_{t-1}\}$.
%\kt{I thought you could not just pre-compute a set of positive passages and subtract selected ones for the next time step; you have to subtract all positive passages that only contain covered answers instead.}
%Specifically, let $\mathcal{P}_t = \{\Tilde{p}_1...\Tilde{p}_t\}$, \sys\ is trained to assign high probabilities to the dynamic oracle given $\mathcal{P}_t$. The objective is defined as follows:\begin{equation*}
%    \sum_{1 \leq t \leq k} \sum_{o \in \mathcal{\Tilde{O}}-\mathcal{P}_{t-1}} - \mathrm{log}P(o|q,\mathcal{B},\mathcal{P}_{t-1}).
%\end{equation*}
At each step $t$ ($1 \leq t \leq k$) , given a set of positive passages $\mathcal{\Tilde{O}}$ and a prefix $\Tilde{p}_1...\Tilde{p}_t$ (denoted as $\mathcal{P}_t$),
\sys\ is trained to assign high probabilities to the dynamic oracle
$\mathcal{\Tilde{O}}-\mathcal{P}_t$. The objective is defined as follows:\begin{equation*}\sum_{1 \leq t \leq k} \sum_{o \in \mathcal{\Tilde{O}}-\mathcal{P}_{t-1}} - \mathrm{log}P(o|q,\mathcal{B},\mathcal{P}_{t-1}).\end{equation*}

%A set of positive passages $\mathcal{\Tilde{O}}$ is obtained at preprocessing time by finding up to $k$ passages with maximal coverage of the distinct answers.\footnote{ $|\mathcal{\Tilde{O}}| < k$ if fewer than $k$ passages are sufficient to cover all distinct answers; $|\mathcal{\Tilde{O}}|=k$ otherwise.} A prefix $\Tilde{p}_1...\Tilde{p}_k$ is then a randomly shuffled combination of positives and negatives so that it can simulate realistic predictions of the model. \hanna{The negative sampling is not clear; what is a baseline? } Specifically, we take $n$ positives from $\mathcal{\Tilde{O}}$ and $k-n$ negatives sampled from $\mathcal{B}-\mathcal{\Tilde{O}}$ based on $s(p_i) + \gamma g_i$, where $s(p_i)$ is a prior logit value from a baseline, $g_i \sim \mathrm{Gumbel}(0, 1)$ and $\gamma$ is a hyperparameter.

\begin{algorithm}[t]
\caption{
    Decoding algorithms for \sys.
    %$k$ is the number of passages to return, $P(p|p_1...p_n)$ is from \sys, and $l$ is a length penalty function.
}\label{alg:decoding}
\begin{algorithmic}[1]
\footnotesize
\Procedure{\naive}{$k, P(p|p_1...p_n)$}
\State $O \gets [ ]$ {\protect\color{gitgreen}  {\textit{// a list of selected passages}}} 
\While{$i=1, \dots, k$}
    \State $\hat{p} \gets \mathrm{argmax}_{p \in \mathcal{B}-O\mathrm{.toSet()}} \mathrm{log}P(p|O)$
    \State $O \gets O::\hat{p}$
\EndWhile
\State \Return $\mathrm{Set}(O)$
\EndProcedure
\Procedure{\decode}{$k, P(p|p_1...p_n), l$}
\State $\mathcal{O} \gets \emptyset$ {\protect\color{gitgreen}  {\textit{// a set of selected passages}}} 
\State $\mathcal{S} \gets [\texttt{Empty}]$ {\protect\color{gitgreen}  {\textit{// a tree}}}
\While{$|\mathcal{O}|<k$}
    \State $P'(p|s) \gets P(p|s) \mathbb{I}[s::p \notin \mathcal{S}]$
    \State $(\hat{s}, \hat{p}) \gets \mathrm{argmax}_{p \in \mathcal{B}, s \in \mathcal{S}} l(|s|+1) \mathrm{log}P'(p|s)$
    %\mathrm{log}P'(p|s)$
    \State $\mathcal{O} \gets \mathcal{O} \cup \{\hat{p}\}, \mathcal{S} \gets \mathcal{S}.\mathrm{append}(\hat{s}::\hat{p})$
\EndWhile
\State \Return $\mathcal{O}$
\EndProcedure
\end{algorithmic}
\end{algorithm}

\subsection{Inference with \decode}\label{subsection:setdecode}
\sm{Updated.}

\hanna{Reorganize the section like this: standard decoding would be (a) SeqDecode that returns the top 1 for each rank in the sequence pushing for diversity of answers, (b) independent decode, that returns all passages from the first step, ignoring dependency between passags. Here, we introduce a TreeDecode that is in between and addresses the trade-off between relevance and diversity. At every step of the generation, it  generate the next item either by expanding the depth or width. 

In particular, we define the score, blah blah...}

A typical autoregressive decoder makes the top 1 prediction at each step to decode a sequence of $k$ (\naive\ in Algorithm~\ref{alg:decoding}),\footnote{We explored beam search decoding but it gives results that are the same as or marginally different from \naive.
} which, based on our training scheme, asks the decoder to find a new answer at every step.
%increases the diversity every step.
However, when $k$ is larger than the number of correct answers, it would be counter-productive to ask for $k$ passages that each covers a distinct answer. Instead, we want the flexibility of decoding fewer timesteps and take multiple predictions from each timestep.
%However, our problem allows the flexibility of taking more than top 1 prediction at a step at the cost of using a smaller number of total steps.
%which helps recovering from errors that may occur in the top 1 prediction.
%Taking multiple predictions from the earlier steps makes the model more robust to early prediction errors.\kl{I see what you mean, but this argument about robustness to early prediction errors is really subtle. I think the previous version talks about controlling the amount of diversity. Do we not believe that this is what is happening? The argument is actually pretty concise and theoretical: during training, each step is trained to explicitly to avoid previous answers, so the naive decoder will be incapable (if trained perfectly) of outputting more passages about the same answers, which is desirable when k > n. A higher level way of describing this (suitable for the abstract) is that there is a train-test mismatch since usually k != n, and TreeDecode is used to fix that discrepancy.}

In this context, we introduce a new decoding algorithm \decode, which decodes a {\em tree} from an autoregressive model. % and returns it as the output set.
\decode\ iteratively chooses between the depth-wise and the width-wise expansion of the tree---going forward to the next step and taking the next best passage within the same step, respectively---until it reaches $k$ passages (Figure~\ref{fig:decoding}).
Intuitively, if the model believes that there are many distinct answers covered by different passages, % and wants to increase the diversity,
it will choose to take the next step, being closer to \naive. On the other hand, if the model believes that there are very few distinct answers, it will choose to take more predictions from the same step, resulting in behavior closer to independent scoring.
%\decode\ allows the model to find a reasonable operating point between \naive\ and independent scoring which are the two extremes.
%Intuitively, if the model believes that there are $k$ distinct answers that need to be covered by the $k$ passages, \decode would always choose to take the next step, reducing it to \naive. In the other extreme, if the model believes that there is only one unique answer to cover, it will take the top $k$ from the first step, equivalent to independent scoring. The truth is typically somewhere between these two extremes and \decode allows the model to choose a reasonable operating point.

The formal algorithm is as follows.
We represent a tree $\mathcal{S}$ as a list of ordered lists $[s_1...s_T]$ where $s_1$ is an empty list and $s_i$ is one element appended to any of $s_1...s_{i-1}$. The corresponding set $\mathcal{O}$ is $\cup_{s \in \mathcal{S}} \mathrm{Set}(s)$.
%Let $\mathrm{Path}(p)$ the path from the root \texttt{null} to the node $p$.
We define a score of a tree $\mathcal{S}$ as
%$$ f(\mathcal{S}) = \sum\limits_{p_1...p_{t_i} \in \mathcal{S}} \sum\limits_{t'=1...t_i} \mathrm{log}P(p_{t'}|p_1...p_{t'-1}).$$
$$ f(\mathcal{S}) = \sum\limits_{p_1...p_{t_i} \in \mathcal{S}} \mathrm{log}P(p_{t_i}|p_1...p_{t_{i-1}}).$$

We form $\mathcal{S}$ and $\mathcal{O}$
through an iterative process by
%by iteratively adding the best addition of an element to $\mathcal{S}$ and $\mathcal{O}$ that maximizes the gain in the score, until we have $k$ unique passages. Specifically, we
(1) starting from $\mathcal{O}=\emptyset$ and $\mathcal{S} = [\texttt{null}]$, and (2) updating $\mathcal{O}$ and $\mathcal{S}$ by finding the best addition of an element that maximizes the gain in $f(\mathcal{S})$, until $|\mathcal{O}|=k$, as described in Algorithm~\ref{alg:decoding}.
%We find $\hat{p}, \hat{s} = \mathrm{argmax}_{p \in \mathcal{B},s \in \mathcal{S}} \mathrm{log}P(p|s)$ under the condition that $\hat{s}::\hat{p}$ is not already in $\mathcal{S}$,\footnote{$::$ denotes concatenation.} and then add $\hat{p}$ and $\hat{s}::\hat{p}$ to $\mathcal{O}$ and $\mathcal{S}$, respectively.

\commentout{
A straightforward way to infer $k$ passages from the model is standard sequence decoding (\naive\ in Algorithm~\ref{alg:decoding}).
Given $P(p|p_1...p_t)$, a probability for each step from the autoregressive model,\footnote{We denote $P(p|q, \mathcal{B}, p_1...p_t)$ as $P(p|p_1...p_t)$ for brevity.} \naive\ takes the top 1 prediction at each step and outputs a sequence of $k$ passages.
This algorithm always looks for a \emph{new} answer at each step.
However in practice, there may be fewer than $k$ distinct answers, and
we want the flexibility of predicting more passages that cover the \emph{same} answer,
which is represented by the next best passage within the same timestep.

In this context, we introduce a new decoding algorithm \decode, which decodes a {\em tree} from an autoregressive model (Algorithm~\ref{alg:decoding}).
We define a scoring function for a tree, and at every step we find the best addition of an element to the set that maximizes the gain in the score, until we have $k$ unique passages.

Formally, we represent a tree $\mathcal{S}$ as a list of ordered lists $[s_1...s_T]$ where $s_1$ is an empty list and $s_i$ is one element appended to any of $s_1...s_{i-1}$.
The corresponding set $\mathcal{O}$ is $\cup_{s \in \mathcal{S}} \text{Set}(s)$.
See Figure~\ref{fig:decoding} for an example.
We define a score of a tree $\mathcal{S}$ as $$ f(\mathcal{S}) = \sum\limits_{p_1...p_{t_i} \in \mathcal{S}} \sum\limits_{t'=1...t_i} \mathrm{log}P(p_{t'}|p_1...p_{t'-1}).$$
We find $\mathcal{S}$ and $\mathcal{O}$ in a greedy manner as described in Algorithm~\ref{alg:decoding}.
The core idea is to (1) keep $\mathcal{O}$ and $\mathcal{S}$ every step starting from $\mathcal{O}=\emptyset$ and $\mathcal{S} = [\texttt{Empty}]$, and (2) update $\mathcal{O}$ and $\mathcal{S}$ by finding the best addition of an element that maximizes the gains in $f(\mathcal{S})$.
We find $\hat{p}, \hat{s} = \mathrm{argmax}_{p \in \mathcal{B},s \in \mathcal{S}} \mathrm{log}P(p|s)$ under the condition that $\hat{s}::\hat{p}$ is not already in $\mathcal{S}$,\footnote{$::$ denotes concatenation.} and then add $\hat{p}$ and $\hat{s}::\hat{p}$ to $\mathcal{O}$ and $\mathcal{S}$, respectively.
An example process is illustrated in Figure~\ref{fig:decoding}.

Intuitively, expanding the depth and the width of the tree represent selecting more passages with and without increasing the diversity, respectively.
If the model believes that there are $k$ distinct answers that need to be covered by the $k$ passages, \decode would always choose to take the next timestep, reducing it to \naive.
In the other extreme, if the model believes that there is only one unique answer to cover, it will take the top $k$ from the first step, equivalent to independent scoring. The truth is typically somewhere between these two extremes and \decode allows the model to choose a reasonable operating point.

To control the trade-off between the depth and the width of the tree, we also use a length penalty function $l$ as shown in Algorithm~\ref{alg:decoding}. Inspired by %the length penalty used in beam search~
\citet{wu2016google}, we use $l(y)=\left(\frac{5+y}{5+1}\right)^\beta$, where $\beta$ is a hyperparameter. }

\section{Experimental Setup}\label{sec:exp-setup}\begin{table}[t]
    \centering \footnotesize
    \begin{tabular}{lrrrrr}
        \toprule
            \multirow{2}{*}{Dataset} &  \multicolumn{3}{c}{\# questions \%} &
            \multicolumn{2}{c}{\# answers}
            \\
            \cmidrule(lr){2-4} \cmidrule(lr){5-6}
            & Train & Dev & Test & Avg. & Median \\
        \midrule
            \webqsp & 2,756 & 241 & 1,582 & 12.4 & 2.0 \\
            \ambigqa & 10,036 & 2,002 & 2,004 & 2.2 & 2.0 \\
            \trec & 1,250 & 119 & 654 & 4.1 & 2.0 \\
        \bottomrule
    \end{tabular}
    \caption{
    Number of questions and an average \& median number of the answers on the \dev\ data.
    Data we use for \trec\ is a subset of the data from \citet{baudivs2015modeling} as described in Appendix~\ref{app:data}.
    %\hanna{make this table as a columnwidth and move it close to the dataset part. -- eg, you can write num of questions to 2.7k, or 10k, etc to save space.}
    }\label{tab:data-statistics}
\end{table}

We compare \sys\ with multiple baselines in a range of multi-answer QA datasets. 
We first present an intrinsic evaluation of passage retrieval by reporting \textsc{MRecall} based on answer coverage in the retrieved passages (Section~\ref{sec:exp}).
We then present an extrinsic evaluation through experiments in downstream question answering (Section~\ref{sec:qa-exp}).

\newcommand{\webqNo}{56.4/37.8}
\newcommand{\webqFlatten}{60.2/40.9}
\newcommand{\webqBase}{60.6/40.2}
    \newcommand{\webqBaseNDCG}{62.4/59.0}
\newcommand{\webqOurs}{\textbf{68.5/56.7}}
    \newcommand{\webqOursNDCG}{\textbf{69.5/67.9}}
\newcommand{\webqNoTen}{61.4/42.5}
\newcommand{\webqFlattenTen}{64.7/45.7}
\newcommand{\webqBaseTen}{65.6/47.2}
    \newcommand{\webqBaseTenNDCG}{60.1/57.2}
\newcommand{\webqOursTen}{\textbf{68.9/55.1}}
    \newcommand{\webqOursTenNDCG}{\textbf{70.3/67.2}}
    
\newcommand{\webqNoT}{57.0/38.9}
\newcommand{\webqFlattenT}{60.2/39.9}
\newcommand{\webqBaseT}{62.9/45.2}
    \newcommand{\webqBaseTNDCG}{65.1/60.9}
\newcommand{\webqOursT}{\textbf{64.9/50.6}}
    \newcommand{\webqOursTNDCG}{\textbf{69.1/65.8}}
\newcommand{\webqNoTenT}{59.0/38.6}
\newcommand{\webqFlattenTenT}{62.9/41.5}
\newcommand{\webqBaseTenT}{63.3/43.1}
    \newcommand{\webqBaseTenTNDCG}{61.0/57.4}
\newcommand{\webqOursTenT}{\textbf{65.7/48.9}}
    \newcommand{\webqOursTenTNDCG}{\textbf{68.9/65.4}}

\newcommand{\ambigNo}{55.2/36.3}
\newcommand{\ambigFlatten}{63.4/43.1}
\newcommand{\ambigBase}{63.7/43.7}
    \newcommand{\ambigBaseNDCG}{73.6/69.5}
\newcommand{\ambigOurs}{\textbf{64.8/45.2}}
    \newcommand{\ambigOursNDCG}{\textbf{73.7/70.0}}
\newcommand{\ambigNoTen}{59.3/39.6}
\newcommand{\ambigFlattenTen}{65.8/46.4}
\newcommand{\ambigBaseTen}{65.5/46.2 }
    \newcommand{\ambigBaseTenNDCG}{73.6/\textbf{69.5}}
\newcommand{\ambigOursTen}{\textbf{67.1/48.2}}
    \newcommand{\ambigOursTenNDCG}{\textbf{73.7}/69.4}
    
\newcommand{\ambigNoT}{-}
\newcommand{\ambigFlattenT}{-}
\newcommand{\ambigBaseT}{-}
    \newcommand{\ambigBaseTNDCG}{-} 
\newcommand{\ambigOursT}{-}
    \newcommand{\ambigOursTNDCG}{-}
\newcommand{\ambigNoTenT}{-}
\newcommand{\ambigFlattenTenT}{-}
\newcommand{\ambigBaseTenT}{-}
    \newcommand{\ambigBaseTenTNDCG}{-}
\newcommand{\ambigOursTenT}{-}
    \newcommand{\ambigOursTenTNDCG}{-}

\newcommand{\trecNo}{53.8/\textbf{29.9}}
\newcommand{\trecFlatten}{53.8/28.4}
\newcommand{\trecBase}{53.8/28.4}
    \newcommand{\trecBaseNDCG}{\textbf{70.7}/61.1}
\newcommand{\trecOurs}{\textbf{55.5/29.9}}
    \newcommand{\trecOursNDCG}{69.8/\textbf{61.4}}
\newcommand{\trecNoTen}{55.5/28.4}
\newcommand{\trecFlattenTen}{55.5/28.4}
\newcommand{\trecBaseTen}{53.8/26.9}
    \newcommand{\trecBaseTenNDCG}{66.4/60.3}
\newcommand{\trecOursTen}{\textbf{56.3/29.9}}
    \newcommand{\trecOursTenNDCG}{\textbf{70.1/62.6}}
    
\newcommand{\trecNoT}{57.8/36.6}
\newcommand{\trecFlattenT}{61.0/39.5}
\newcommand{\trecBaseT}{61.9/39.2}
    \newcommand{\trecBaseTNDCG}{\textbf{74.9}/66.4}
\newcommand{\trecOursT}{\textbf{62.4/41.1}}
    \newcommand{\trecOursTNDCG}{74.7/\textbf{66.8}}
\newcommand{\trecNoTenT}{60.1/38.4}
\newcommand{\trecFlattenTenT}{\textbf{64.8}/43.0}
\newcommand{\trecBaseTenT}{{63.8}/42.2}
    \newcommand{\trecBaseTenTNDCG}{68.9/61.5}
\newcommand{\trecOursTenT}{64.5/\textbf{43.3}}
    \newcommand{\trecOursTenTNDCG}{\textbf{74.3/66.2}}

\newcommand{\webqQANoLarge}{51.4/47.0}
\newcommand{\webqQANo}{50.7/45.3}
\newcommand{\webqQABase}{51.8/46.9}
\newcommand{\webqQAOurs}{\textbf{53.6/49.5}}

\newcommand{\webqQANoLargeT}{52.4/45.8}
\newcommand{\webqQANoT}{51.9/45.0}
\newcommand{\webqQABaseT}{51.8/45.0}
\newcommand{\webqQAOursT}{\textbf{53.1/47.2}}

\newcommand{\ambigQANoLarge}{45.5/34.9}
\newcommand{\ambigQANo}{ 43.5/34.6 }
\newcommand{\ambigQABase}{47.6/36.2 }
\newcommand{\ambigQAOurs}{\textbf{48.5/37.6}}
\newcommand{\ambigQASOTA}{48.3/37.3}

\newcommand{\ambigQANoLargeT}{41.1/30.9}
\newcommand{\ambigQANoT}{ 39.6/31.4	 }
\newcommand{\ambigQABaseT}{42.3/32.0}
\newcommand{\ambigQAOursT}{\textbf{43.5/34.2}}
\newcommand{\ambigQASOTAT}{42.1/33.3}

\newcommand{\trecQANoLarge}{40.1/32.8}
\newcommand{\trecQANo}{ 46.2/32.2 }
\newcommand{\trecQABase}{44.6/32.8}
\newcommand{\trecQAOurs}{\textbf{48.6/32.8}}

\newcommand{\trecQANoLargeT}{42.5/32.2}
\newcommand{\trecQANoT}{44.7/32.1}
\newcommand{\trecQABaseT}{45.9/31.8}
\newcommand{\trecQAOursT}{\textbf{46.8/33.3}}

\begin{table*}[t]
    \centering \footnotesize
    \begin{tabular}{l p{4cm} P{1.3cm} P{1.3cm} P{1.3cm} P{1.3cm} P{1.3cm}}
        \toprule
            \multirow{2}{*}{$k$} & \multirow{2}{*}{Models} & \multicolumn{2}{c}{\webqsp} &
            \multicolumn{1}{c}{\ambigqa} &
            \multicolumn{2}{c}{\trec} \\
            \cmidrule(lr){3-4} \cmidrule(lr){5-5} \cmidrule(lr){6-7}
            && Dev & Test & Dev & Dev & Test \\
        \midrule
            %\bolden{$k=5$} \\
            \multirow{4}{*}{$5$} & \norerank & \webqNo & \webqNoT & \ambigNo & \trecNo & \trecNoT \\
            & \nogueira & \webqFlatten & \webqFlattenT & \ambigFlatten & \trecFlatten & \trecFlattenT \\
            & \baseline & \webqBase & \webqBaseT & \ambigBase & \trecBase & \trecOursT  \\
            & \ours & \webqOurs & \webqOursT & \ambigOurs & \trecOurs & \trecOursT \\
        \midrule
            %\bolden{$k=10$} \\
            \multirow{4}{*}{$10$} & \norerank & \webqNoTen & \webqNoTenT & \ambigNoTen & \trecNoTen & \trecNoTenT \\
            & \nogueira & \webqFlattenTen & \webqFlattenTenT & \ambigFlattenTen & \trecFlattenTen & \trecFlattenTenT \\
            & \baseline & \webqBaseTen & \webqBaseTenT  & \ambigBaseTen & \trecBaseTen & \trecBaseTenT \\
            & \ours & \webqOursTen & \webqOursTenT & \ambigOursTen  & \trecOursTen & \trecOursTenT  \\
        \bottomrule
    \end{tabular}
    \caption{
    Results on passage retrieval in \textsc{MRecall}. %\mc{use the MRecall}
    The two numbers in each cell indicate performance on all questions and on questions with more than one answer, respectively.
    Test-set metrics on \ambigqa\ are not available as its test set is hidden, but we report the test results on question answering in Section~\ref{sec:qa-exp}. \\
    \footnotesize{Note: it is possible to have higher \textsc{MRecall} @ 5 than \textsc{MRecall} @ 10 based on our definition of \textsc{MRecall} (Section~\ref{subsec:multi-answer-retrieval}). %, e.g., a case with six distinct answers of which five are covered by the top 1--5 passages but none covered by the top 6--10.
    }
    }\label{tab:results}
\end{table*}

\subsection{Datasets}\label{subsec:data}
We train and evaluate on three datasets that provide a set of distinct answers for each question. Statistics of each dataset are provided in Table~\ref{tab:data-statistics}.

\noindent \textbf{\webqsp}~\citep{yih2016value} consists of questions from Google Suggest API, originally from \citet{berant2013semantic}. The answer is a set of distinct entities in Freebase; we recast this problem as textual question answering %and select the answer text from
based on Wikipedia. %\footnote{Data splits from \url{github.com/OceanskySun/GraftNet\#downloads}.}

\noindent \textbf{\ambigqa}~\citep{min2020ambigqa} consists of questions mined from Google search queries, originally from \nq~\citep{kwiatkowski2019natural}. Each question is paired with an annotated set of distinct answers that are equally valid based on Wikipedia.

\noindent \textbf{\trec}~\citep{baudivs2015modeling} contains questions curated from TREC QA tracks, along with regular expressions as answers. 
Prior work uses this data as a task of finding a single answer (where retrieving any of the correct answers is sufficient), but we recast the problem as a task of finding all answers, and approximate a set of distinct answers. Details are described in Appendix~\ref{app:data}.

\subsection{First-stage Retrieval}\label{subsec:dense-retrieval}
\sys\ can obtain candidate passages $\mathcal{B}$ from any first-stage retrieval model. In this paper, we use \dpr, our own improved version of DPR~\citep{karpukhin2020dense} combined with REALM~\citep{guu2020realm}.
DPR and REALM are dual encoders with a supervised objective and an unsupervised, language modeling objective, respectively.
We initialize the dual encoder with REALM and train on supervised datasets using the objective from DPR.
More details are provided in Appendix~\ref{app:dense-retrieval}.
%\hanna{DPR+ is not clear; it is not enought it is explained only in appendix}
%\sm{I describe \dpr\ here but IMO it's not the best place to describe it because this section is all about baselines but \dpr\ is also used in \sys.}

\begin{table*}[t]
    \parbox{.38\linewidth}{\centering \footnotesize
    \begin{tabular}{p{3.7cm} P{1.5cm} }
        \toprule
            Training method & \textsc{MRecall} \\
        \midrule
            Dynamic oracle & \textbf{67.6/56.7} \\
            Dynamic oracle w/o negatives & 65.1/52.0 \\
            Teacher forcing & 66.4/51.2 \\
        \bottomrule
    \end{tabular}
    \caption{
    Ablations in training methods for \sys.
    Results on \webqsp\ ($k=5$).
    All rows use \naive\ (instead of \decode).
    \hanna{In Table 4, you show that adding negatives is really important and only positive with your training is worse than teacher forcing. you can do this negative mining because of your dynamic oracle training?}
    }\label{tab:ablations-prefix}
    }
    \hfill
    \parbox{.58\linewidth}{\centering \footnotesize
    \begin{tabular}{l p{2.2cm} P{0.4cm} P{1.5cm} P{0.4cm} P{1.5cm}}
        \toprule
            \multirow{2}{*}{$k$} & \multirow{2}{*}{Decoding} &  \multicolumn{2}{c}{\webqsp} &
            \multicolumn{2}{c}{\ambigqa} \\
            \cmidrule(lr){3-4} \cmidrule(lr){5-6}
            && $d$ & \textsc{MRecall} & $d$ & \textsc{MRecall}  \\
        \midrule
            \multirow{2}{*}{$5$}
            & \textsc{\naive} & 5.0 & 67.6/\textbf{56.7} & 5.0 & 63.1/42.5 \\
            & \decode & 3.0 & \webqOurs & 2.1 & \ambigOurs \\
        \midrule
            \multirow{2}{*}{$10$}
            & \textsc{\naive} & 10.0 & 68.0/54.3 & 10.0 & 65.0/45.9 \\
            & \decode & 5.4 & \webqOursTen & 2.9 & \ambigOursTen \\
        \bottomrule
    \end{tabular}
    \caption{Ablations in decoding methods for \sys. $d$ refers to the average depth of the tree ($\mathrm{max}_{s \in \mathcal{S}} |s|$ in Algorithm~\ref{alg:decoding}).    
    }\label{tab:ablations-decode}
    }
\end{table*}

\begin{table*}[t]
    \centering \footnotesize %\scriptsize \setlength\tabcolsep{4pt}
    \begin{tabular}{l | l}
        \toprule
            \multicolumn{2}{c}{Q: Who play Mark on the TV show Roseanne?} \\
        \midrule
            \base & \sys \\
            \textbf{\#1} \hl{Glenn Quinn} ... He was best known for his portrayal of & \textbf{\#1} \hl{Glenn Quinn} ... He was best known for his portrayal of \\
            Mark Healy on the popular '90s family sitcom Roseanne. & Mark Healy on the popular '90s family sitcom Roseanne. \\
            \textbf{\#2} 
            \hl{Glenn Quinn}, who played Becky's husband, Mark, died & \textbf{\#2} Becky begins dating Mark Healy (\hl{Glenn Quinn}) ... \\
            in December 2002 of a heroin overdose at the age of 32 ... & \textbf{\#3} 
            \hl{Glenn Quinn}, who played Becky's husband, Mark, died\\
            \textbf{\#3} Becky begins dating Mark Healy (\hl{Glenn Quinn}) ... & in December 2002 of a heroin overdose at the age of 32 ... \\
            \textbf{\#4} Johnny Galecki ...  on the hit ABC sitcom Roseanne as & \textbf{\#4} Roseanne (season 10) ... In September 2017, \hl{Ames}  \\
            the younger brother of Mark Healy (\hl{Glenn Quinn}) ... &
            \hl{McNamara} was announced to be cast as Mark Conner-Healy. \\
        \bottomrule
    \end{tabular}
    \caption{
    An example prediction from \base\ and \sys; answers to the input question highlighted.
    While \base\ repeatedly retrieves passages supporting the same answer {\em Glenn Quinn} and fails to cover other answers, \sys\ successfully retrieves a passage covering a novel answer {\em Ames McNamara}.
    }\label{tab:example}
\end{table*}

\subsection{Baselines}\label{subsec:baseline}
We compare \sys\ with three baselines,
all of which are published models or enhanced versions of them.
All baselines independently score each passage.

\vspace{.1cm}
\noindent \textbf{\norerank} uses \dpr\ without a reranker. 

\vspace{.1cm}
\noindent \textbf{\nogueira} uses \dpr\ followed by \citet{nogueira-etal-2020-document}, the \sota\ document ranker.
It processes each passage $p_i$ in $\mathcal{B}$ independently and is trained to output \texttt{yes} if $p_i$ contains any valid answer to $q$, otherwise \texttt{no}.
%The length of the decoded sequence is always 1.
At inference, the probability for each $p_i$ is computed by taking a softmax over the logit of \texttt{yes} and \texttt{no}. The top $k$ passages are chosen based on the probabilities assigned to \texttt{yes}.

\vspace{.1cm}
\noindent \textbf{\base} is our own baseline that is a strict non-autoregressive version of \sys\ in which prediction of a passage is independent from other passages in the retrieved set.
It obtains candidate passages $\mathcal{B}$ through \dpr\ and the encoder of the reranker processes $q$ and $\mathcal{B}$, as \sys\ does.
Different from \sys, the decoder is trained to output a single token $i$ ($1 \leq i \leq |\mathcal{B}|$) rather than a sequence.
The objective is the sum of $-\mathrm{log}P(p|q,\mathcal{B})$ of the passages including any valid answer to $q$.
At inference, \base\ outputs the top $k$ passages based the logit values of the passage indices.
We compare mainly to \base\ because it is the strict non-autoregressive version of \sys, and is empirically better than or comparable to \citet{nogueira-etal-2020-document} (Section~\ref{sec:exp}).

%\vspace{.1cm}
%\noindent \textbf{\base} is our own baseline that is a strict non-autoregressive version of \sys. It is the same as \sys\ in that it obtains candidate passages $\mathcal{B}$ through dense retrieval and the encoder of the reranker processes $q$ and $\mathcal{B}$. Different from \sys, the decoder is trained to output a single token $i$ ($1 \leq i \leq |\mathcal{B}|$) rather than a sequence. \hanna{single token for what? a single token i for a given pi?, and hence lightly encodes interactions between passages?}
%The objective is the sum of $-\mathrm{log}P(p|q,\mathcal{B})$ of the passages including any valid answer to $q$.
%\kt{I thought -log (sum) worked significantly better at one point.} \sm{We actually found sum of log works significantly better.}
%At inference, we output the top $k$ passages based the logit values of the passage indices.
%We compare mainly to \base because it is conceptually close to the state-of-the-art \nogueira\footnote{\base adds minor improvements through (1) lightweight interactions between candidate passages in the decoding step, \hanna{the difference is generating i instead of yes/no, and that adds interactions? }\sm{yes!} and (2) a ranking loss instead of a binary loss.}, while the implementation is a minimal ablation of the autoregressive modeling in \sys.

\subsection{Implementation Details}\label{subsec:impl-details}

We use the English Wikipedia from 12/20/2018 as the retrieval corpus $\mathcal{C}$, where each article is split into passages with up to 288 wordpieces,
%\hanna{next sentence is not clear; what is DPR+, why do you call it DPR+, what additions does it have over DPR. I think you should introduce DPR+ in the baseline section and then remove it from here.}
%\dpr\ is a combination of recent advances in dense retrieval, based on REALM~\citep{guu2020realm} followed by a supervised training objective~\citep{karpukhin2020dense}. More details are provided in Appendix~\ref{app:dense-retrieval}.
%
All rerankers are based on T5~\citep{raffel2020exploring}, a pretrained encoder-decoder model; %, inspired by recent studies that showed the importance of using large-scale pre-trained models~\citep{devlin2019bert,raffel2020exploring}.
T5-base is used unless otherwise specified.
We use $|\mathcal{B}|=100, k=\{5, 10\}$.
Models are first trained on \nq~\citep{kwiatkowski2019natural} and then finetuned on multi-answer datasets, which we find helpful since all multi-answer datasets are relatively small.
During dynamic oracle training, $k-|\mathcal{\Tilde{O}}|$ negatives are sampled from $\mathcal{B}-\mathcal{\Tilde{O}}$ based on $s(p_i) + \gamma g_i$, where $s(p_i)$ is a prior logit value from \base, $g_i \sim \mathrm{Gumbel}(0, 1)$ and $\gamma$ is a hyperparameter.
In \decode, to control the trade-off between the depth and the width of the tree, we use a length penalty function $l(y)=\left(\frac{5+y}{5+1}\right)^\beta,$ where $\beta$ is a hyperparameter, following \citet{wu2016google}. \hanna{not clear about the l(y)}
More details are in Appendix~\ref{app:training-details}.

\section{Experimental Results}
\subsection{Retrieval Experiments}\label{sec:exp}Table~\ref{tab:results} reports \textsc{MRecall} on all questions and on questions with more than one answer.

\vspace{-.3em}
\noindent \paragraph{No reranking vs. reranking}
Models with reranking (\nogueira, \base\ or \sys) are always better than \norerank, demonstrating the importance of reranking.

\vspace{-.3em}
\noindent \paragraph{Independent vs. joint ranking}
\sys\ consistently outperforms both \nogueira\ and \base on all datasets and all values of $k$.
Gains are especially significant on questions with more than one answer, outperforming two reranking baselines by up to 11\% absolute and up to 6\% absolute, respectively.
\webqsp\ sees the largest gains out of the three datasets, likely because the average number of answers is the largest. %\hanna{for example, here add, showing more improvements in questions with multiple answers.} \sm{Added in the sentence before this one.}

\subsubsection{Ablations \& Analysis}\label{subsec:ablation}

\paragraph{Training methods}
Table~\ref{tab:ablations-prefix} compares dynamic oracle training with alternatives.
`Dynamic oracle w/o negatives' is the same as dynamic oracle training except the prefix only has positive passages. %, without negative passages.
`Teacher forcing' is a standard method in training an autoregressive model:
given a target sequence $o_1...o_k$, the model is trained to maximize $\Pi_{1\leq t \leq k}P(o_t|o_1...o_{t-1})$.
We form a target sequence using a set of positive passages $\Tilde{\mathcal{O}}$, where the order is determined by following the ranking from \base.
%``Positives only" forms a prefix with positive passages only.
Table~\ref{tab:ablations-prefix} shows that our dynamic oracle training, which uses both positives and negatives, significantly outperforms the other methods. %\kt{Can clarify that setdecode is used in both cases for inference}

\vspace{-.3em}
\paragraph{Impact of \decode}
Table~\ref{tab:ablations-decode} compares \sys\ with \naive\ and with \decode. We find that \decode\ consistently improves the performance on both \webqsp\ and \ambigqa, with both $k=5$ and $10$.
Gains are especially significant on \ambigqa, since the choice of whether to increase diversity is more challenging on \ambigqa\, where questions are more specific and have fewer distinct answers, which \decode\ better handles compared to \naive.
%: we think this is because \decode\ can better control the trade-off between whether to increase diversity or not \kt{I don't understand this point well}, and it matters more for \ambigqa\ with more specific questions and a smaller number of distinct answers.
The average depth of the tree
%($\mathrm{max}_{s \in \mathcal{S}} |s|$ in Algorithm~\ref{alg:decoding}, denoted as $d$ in Table~\ref{tab:ablations-decode})
is larger on \webqsp, likely because its average number of distinct answers is larger and thus requires more diversity.

\begin{table*}[t]
    \centering \footnotesize
    \begin{tabular}{l l l l P{1.1cm} P{1.1cm} P{1.1cm} P{1.1cm} P{1.1cm} P{1.1cm}}
        \toprule
            \multirow{2}{*}{Retrieval} & 
            \multirow{2}{*}{QA Model} &
            \multirow{2}{*}{$k$} & \multirow{2}{*}{Mem} &
            \multicolumn{2}{c}{\webqsp} &
            \multicolumn{2}{c}{\ambigqa} &
            \multicolumn{2}{c}{\trec} \\
            \cmidrule(lr){5-6} \cmidrule(lr){7-8} \cmidrule(lr){9-10}
            &&&& Dev & Test & Dev & Test & Dev & Test  \\
        \midrule
            \norerank    & T5-3B & 10 & x1
            & \webqQANo & \webqQANoT & \ambigQANo & \ambigQANoT & \trecQANo & \trecQANoT \\
            \baseline      & T5-3B & 10 & x1
            & \webqQABase & \webqQABaseT & \ambigQABase & \ambigQABaseT & \trecQABase & \trecQABaseT \\
            \ours            & T5-3B & 10 & x1
            & \webqQAOurs & \webqQAOursT & \ambigQAOurs & \ambigQAOursT & \trecQAOurs & \trecQAOursT \\
        \midrule
            \norerank    & T5-large & 40 & x1
            & \webqQANoLarge & \webqQANoLargeT & \ambigQANoLarge & \ambigQANoLargeT & \trecQANoLarge & \trecQANoLargeT \\
        \midrule
            %\multicolumn{3}{l}{\citet{gao2020answering}}
            \citet{gao2020answering} & BART-large & 100 & x3
            & - & - & \ambigQASOTA & \ambigQASOTAT & - & - \\
        \bottomrule
    \end{tabular}
    \caption{
    Question Answering results on multi-answer datasets.
    The two values in each cell indicate F1 on all questions and F1 on questions with multiple answers only, respectively.
    {\em Mem} compares the required hardware memory during training.
    Note that \citet{gao2020answering} reranks 1000 passages instead of 100, and trains an answer generation model using x3 more memory than ours. {\bf Better retrieval enables using larger answer generation models on fewer retrieved passages.}
    }\label{tab:qa-results}
\end{table*}

\vspace{-.3em}
\paragraph{An example prediction}
Table~\ref{tab:example} shows predictions from \base\ and \sys\ given an example question from \ambigqa, ``Who plays Mark on the TV show Roseanne?'' One answer {\em Glenn Quinn} is easy to retrieve because there are many passages in Wikipedia providing evidence, %mentioning that he played Mark on Roseanne,
while the other answer {\em Ames McNamara} is harder to find. While \base\ repeatedly retrieves passages that mention {\em Glenn Quinn} and fails to cover {\em Ames McNamara}, \sys\ successfully retrieves both answers.

More analysis can be found in Appendix~\ref{app:additional-results}.

\subsection{QA Experiments}\label{sec:qa-exp}%This section discusses experiments on downstream question answering: given a question and a set of passages from retrieval,
%the model should output all valid answers to the question.
%
%We use an answer generation model based on~\citet{izacard2020leveraging}
%which we train to generate a sequence of answers, separated by a \texttt{[SEP]} token.
This section discusses experiments on downstream question answering: given a question and a set of
passages from retrieval, the model outputs
all valid answers to the question. 
We aim to answer two research questions: (1) whether the improvements in passage retrieval are transferred to improvements in downstream question answering, and (2) whether using a smaller number of passages through reranking is  better than using the largest possible number of passages given fixed hardware memory. 

We use an answer generation model based on~\citet{izacard2020leveraging}
which we train to generate a sequence of answers, separated by a \texttt{[SEP]} token, given a set of retrieved passages.
%\paragraph{Evaluation Metrics}\label{subsec:qa-metric}
%For multi-answer datasets, we follow \citet{min2020ambigqa} and use F1, which computes an F1 score between the gold set of answers and the predicted set of answers.
%We report F1 scores on all questions and on questions with more than one answer.
%For \nq, we follow prior work in using Exact Match accuracy (EM).
Our main model uses \sys\ to obtain passages fed into the answer generation model. The baselines obtain passages from either \textbf{\norerank} or \textbf{\base}, described in Section~\ref{subsec:baseline}.

\newcommand{\mybase}{$\{k=140,$ base$\}$}
\newcommand{\mylarge}{$\{k=40,$ large$\}$}
\newcommand{\mythreeb}{$\{k=10,$ 3B$\}$}

We compare different models that fit on the same hardware 
by varying the sizes of T5 (base, large, 3B) and use the maximum number of passages ($k$).\footnote{The memory requirement is $O(k \times \text{T5 size})$.}
This results in three settings: \mybase, \mylarge\ and \mythreeb.

%We report F1 scores for multi-answer datasets and report Exact Match for \nq, following \citet{min2020ambigqa} and \citet{chen2017reading}, respectively.

\subsubsection{Main Result}\label{subsec:qa-result}
Table~\ref{tab:qa-results} reports the performance on three multi-answer datasets in F1, following \citet{min2020ambigqa}.

\vspace{-.3em}
\paragraph{Impact of reranking}
With \mythreeb, \sys\ outperforms both baselines, indicating that the improvements in retrieval are successfully transferred to improvements in QA performance.
We however find that our sequence-to-sequence answer generation model tends to undergenerate answers, presumably due to high variance in the length of the output sequence.
This indicates the model is not fully benefiting from retrieval of many answers, and we expect more improvements when combined with an answer generation model that is capable of generating many answers.

\vspace{-.3em}
\paragraph{More passages vs. bigger model}
With fixed memory during training, using fewer passages equipped with a larger answer generation model outperforms using more passages.
%we compare with a model that uses more passages and a smaller answer generation model (\mylarge), and find it significantly worse than \sys\ with fewer passages and a bigger answer generation model.
This is only true when reranking is used; otherwise, using more passages is often better or comparable.
%This demonstrates that reranking provides merits of using fewer passages and a larger answer generation model, ultimately leading to the best performance.
This demonstrates that, as retrieval improves, memory is better spent on larger answer generators rather than more passages, leading to the best performance.

\vspace{.3em}
Finally, \sys\ establishes a new \sota, outperforming the previous \sota\ on \ambigqa~\citep{gao2020answering} with extensive reranking and the answer generation model trained using x3 more resources than ours.\footnote{\citet{gao2020answering} reranks 1000 passages through independent scoring as in \citet{nogueira-etal-2020-document}; it is not a directly comparable baseline and serves as a point of reference.}

\begin{table}[t]
    \centering \footnotesize
    \begin{tabular}{l l l P{1cm} P{1cm}}
        \toprule
            Model & T5 & $k$ & dev & test \\
        \midrule
            \norerank    & base & 140 & 46.4 & - \\
            \norerank    & large & 40 & 47.3 & - \\
            \norerank    & 3B & 10 & 46.5 & - \\
        \midrule
            \baseline     & large & 40 & 49.4 & -\\
            \baseline     & 3B & 10 & 50.4 & \textbf{54.5} \\
        \midrule
            %\multicolumn{3}{l}{\citet{karpukhin2020dense}} & - & 41.5 \\
            %\multicolumn{3}{l}{\citet{izacard2020leveraging} base} & - & 48.2 \\
            \multicolumn{3}{l}{\citet{izacard2020leveraging} %large
            } & - & 51.4 \\
        \bottomrule
    \end{tabular}
    \caption{
    Question Answering results on \nq.
    We report Exact Match (EM) accuracy.
    The first five rows are from our own experiments, which all use the same hardware resources for training.
    %The last three rows are the published and unpublished state of the art.
    %Note that \citet{izacard2020leveraging} uses x1.4 and x5 more resources than ours to train the base and the large versions of their model, respectively.
    %The last two rows are the previous \sota, which use x1.4 and x5 more resources than ours to train the base and the large versions of their model, respectively.
    The last row is the previous \sota\ which requires x5 more resources than ours to train the model.
    %This result justifies our choice of focusing on reranking, and shows that our baseline is already beating the competitive \sota\ on \nq.
    }\label{tab:nq-qa-results}
\end{table}

\vspace{-.3em}
%\paragraph{Single-answer QA result on \nq}
\subsubsection{Single-answer QA result}
While our main contributions are in multi-answer retrieval, we experiment on \nq\ to demonstrate that the value of good reranking extends to the single-answer scenario.
Table~\ref{tab:nq-qa-results} indicates two observations consistent to the findings from multi-answer retrieval:
(1) when compared within the same setting (same T5 and $k$), \base\ always outperforms \norerank, %demonstrating that passages with higher quality indeed lead to improvements in QA performance.
and
(2) with reranking, \mythreeb\ %with fewer passages 
outperforms \mylarge. %\ with more passages.
%Next, when compared between models with reranking, \mythreeb\ outperforms \mylarge. This is only true when reranking is used; \mythreeb\ does not outperform \mylarge\ without reranking, likely because the quality of passages is insufficient when $k$ is small. This demonstrates that higher quality passages through reranking improve the performance by virtue of using fewer passages and a larger answer generation model.
%
%In Table~\ref{tab:nq-qa-results},
Finally, our best model outperforms the previous \sota~\citep{izacard2020leveraging} which uses x5 more training resources.
Altogether, this result (1) justifies our choice of focusing on reranking, and (2) shows that \base\ is very competitive and thus our \sys\ results in multi-answer retrieval are very strong.

\section{Related Work}\label{sec:related}We refer to Section~\ref{sec:task} for related work focusing on single-answer retrieval.

%\paragraph{Passage retrieval}
%\hanna{you can remove this paragraph from here; then move things to appropriate places (either in the intro) or the background section). } Retrieval is crucial both as an end task for users~\citep{jarvelin2002cumulated} and for downstream tasks such as question answering~\citep{chen2017reading} or generation~\citep{rag}. It has extensively been studied for decades, where the best practice is to use an efficient method~\citep{ramos2003using,robertson2009probabilistic,yih2011learning,lee2019latent,karpukhin2020dense,luan2020sparse} followed by a more expensive reranker~\citep{liu2011learning,asadi2013effectiveness,nogueira2019passage,nogueira-etal-2020-document}. Nearly all work aims to retrieve passages that only need to contain at least one answer. Existing models score each passage independently and output the top $k$ to approximate set retrieval.

%In this paper, we tackle the underexplored yet important problem of finding all valid answers, requiring joint modeling of a set of passages.

\vspace{-.3em}
\paragraph{Diverse retrieval}
Studies on diverse retrieval in the context of information retrieval (IR) requires finding documents covering many different sub-topics to a query topic~\citep{zhai2003beyond,clarke2008novelty}. 
Questions are typically underspecified, and many documents (e.g. up to 56 in \citet{zhai2003beyond}) are considered relevant.
In their problem space, effective models post-hoc increase the distances between output passages during inference~\citep{zhai2003beyond,abdool2020managing}.

Our problem is closely related to diverse retrieval in IR, with two important differences.
First, since questions represent more specific information needs, controlling the trade-off between relevance and diversity is harder, and simply increasing the distances between retrieved passages does not help.\footnote{
    In our preliminary experiment, we tried increasing diversity based on Maximal Marginal Relevance~\citep{carbonell1998use} following~\citet{zhai2003beyond,abdool2020managing}; it improves diversity but significantly hurts the relevance to the input question, dropping the overall performance.
}
Second, multi-answer retrieval uses a clear notion of ``answers''; ``sub-topics'' in diverse IR are more subjective and hard to enumerate fully.% in diverse retrieval are often vaguely defined.

\vspace{-.3em}
\paragraph{Multi-hop passage retrieval}
Recent work studies multi-hop passage retrieval, where 
a passage containing the answer is the destination of a chain of multiple hops~\citep{asai2019learning,xiong2021answering,khattab2021baleen}.
This is a difficult problem as passages in a chain are dissimilar to each other, but existing datasets often suffer from annotation artifacts~\citep{chen2019understanding,min2019compositional}, resulting in strong lexical cues for each hop. We study an orthogonal problem of finding multiple answers, where the challenge is in controlling the trade-off between relevance and diversity.

\section{Conclusion}\label{sec:concl}We introduce \sys, an autoregressive passage reranker designed to address the multi-answer retrieval problem.
On three multi-answer datasets, \sys\ significantly outperforms a range of baselines in both retrieval recall and downstream QA accuracy, establishing a new \sota.
Future work could extend the scope of the problem to other tasks that exhibit specific information need while requiring diversity.

\section*{Acknowledgements}
We thank the Google AI Language members, the UW NLP members, and the anonymous reviewers for their valuable feedback.
This work was supported in part by ONR N00014-18-1-2826 and DARPA N66001-19-2-403.

\bibliographystyle{acl_natbib}
\bibliography{emnlp2021}

\clearpage
\appendix
\section{Details of \dpr}\label{app:dense-retrieval}
We use a pretrained dual encoder model from REALM~\citep{guu2020realm} and further finetune it on the QA datasets using the objective from DPR~\citep{karpukhin2020dense}:
\begin{equation*}
    \mathcal{L} = -\mathrm{log} \frac{f_q(q)^T f_p(p^\mathrm{+})}{\sum_{p \in \{p^\mathrm{+}\} \cup \mathcal{B}^\mathrm{-}} f_q(q)^T f_p(p)},
\end{equation*}
where $f_q$ and $f_p$ are trainable encoders for the questions and passages, respectively, $p^\mathrm{+}$ is a positive passage (i.e., a passage containing the answer), and $\mathcal{B}^\mathrm{-}$ is a set of negative passages (i.e., passages without the answer).
As shown in~\citet{karpukhin2020dense}, a choice of $\mathcal{B}^\mathrm{-}$ is significant for the performance. We explore two methods:

\noindent \textbf{Distant negatives} follows DPR~\citep{karpukhin2020dense} in using distantly obtained negative passages as $\mathcal{B}^\mathrm{-}$.
We obtain two distant negative passages per question:
one hard negative, a top prediction from REALM without finetuning, and one random negative, drawn from a uniform distribution, both not containing the answer.

\noindent \textbf{Full negatives} considers all passages in Wikipedia expect $p^\mathrm{+}$ as $\mathcal{B}^\mathrm{-}$, and instead freezes the passage encoder $f_p$ and only finetunes the question encoder $f_q$.
This is appealing because (a) the number and the quality of the negatives, which both are the significant factors for training, are the strict maximum, and (b) $f_p$ from REALM is already good, producing high quality passage representations without finetuning. Implementation of this method is feasible by exploiting extensive model parallelism.

We use distant negatives for multi-answer datasets and full negatives for \nq\ as this combination gave the best result.

\commentout{
\section{Explanations of \decode}\label{app:setdecode}
This section continues from Section~\ref{subsection:setdecode} and explains how \decode\ in Algorithm~\ref{alg:decoding} is equivalent to finding the best addition of the passage in the tree that maximizes the score of the tree.

To recap, a score of a tree $\mathcal{S}$ is defined as $$ f(\mathcal{S}) = \sum\limits_{p_1...p_{t_i} \in \mathcal{S}} \sum\limits_{t'=1...t_i} \mathrm{log}P(p_{t'}|p_1...p_{t'-1}).$$
The goal is, given a current tree $\mathcal{S}$, find the passage $p$ that maximizes the gains in the score: \begin{eqnarray*}
    g(p|\mathcal{S}) &=& \mathrm{max}_{s \in S} f(\mathcal{S} \cup \{s::p\}) - f(\mathcal{S}).
\end{eqnarray*}
Based on the definition of $f$,
\begin{eqnarray*}
    f(\mathcal{S} \cup \{s::p\}) = \begin{cases}
        f(\mathcal{S}) & \text{if } s::p \in \mathcal{S}, \\
        f(\mathcal{S}) + \mathrm{log}P(p|s) & \text{otherwise}.
    \end{cases}
\end{eqnarray*}
Therefore, 
\begin{eqnarray*}
    g(p|\mathcal{S}) &=& \mathrm{max}_{s \in S} \mathrm{log}P(p|s) \mathbb{I}[s::p \notin \mathcal{S}]
\end{eqnarray*}
Finally, an optimal addition of the passage $\hat{p}$ to an existing branch in the tree $s$ is computed via:
$$\hat{p}, \hat{s} = \mathrm{argmax}_{p \in \mathcal{B},s \in \mathcal{S}} \mathrm{log}P(p|s) \mathbb{I}[s::p \notin \mathcal{S}].$$
The addition of $\hat{p}$ updates $\mathcal{O}$ and $\mathcal{S}$ by adding $\hat{p}$ and $[\hat{s}::\hat{p}]$, respectively. This process is equivalent to L11--13 in Algorithm~\ref{alg:decoding}.
}

\section{Experiment Details}

\subsection{Data processing for \trec}\label{app:data}

\trec\ from \citet{baudivs2015modeling} contains regular expressions as the answers. We approximate a set of semantically distinct answers as follows.
We first run regular expressions over Wikipedia to detect valid answer text.
If there is no valid answer found from Wikipedia, or there are more than 100 valid answers\footnote{In most of such cases, the regular expressions are extremely permissive.}, we discard the question.
We then only keep the answers with up to five tokens, following the notion of short answers from \citet{lee2019latent}.
Finally, we group the answers that are the same after normalization and white space removal.
We find that this gives a reasonable approximation of a set of semantically distinct answers.
Note that the data we use is the subset of the original data because we discarded a few questions. Statistics are reported in Section~\ref{subsec:data}.

Here is an example: a regular expression from the original data is \texttt{Long Island|New\textbackslash s?York|Roosevelt Field}. All matching answers over Wikipedia include \texttt{roosevelt field}, \texttt{new york}, \texttt{new\textbackslash xa0york}, \texttt{new\textbackslash nyork}, \texttt{newyork}, \texttt{long island}.
Once the grouping is done, we have three semantically distinct answers: (1) \texttt{roosevelt field}, (2) \texttt {new york|new\textbackslash xa0york|new\textbackslash nyork|newyork}, and (3) \texttt{long island}.

%\subsection{Details on adding noise in training examples for QA}\label{app:training-qa-noise}
%In order to combat the problem of overfitting in the pipeline model, especially for training the QA model, we add noise in the retrieval predictions on training data, to match the recall accuracy between the training and \dev\ data. In particular, given an original, normalized logit value $z(b_i)$ for each element (passage) $b_i$, a new logit value is defined as $\hat{z}(b_i) = z(b_i) + \lambda \mathrm{gumbel\_noise}()$, where $\lambda$ is a hyperparameter.

%For \sys, assigning a logit value for each $b_i$ is non-trivial. We therefore approximate the logit by $z(b_i) = k / (r_i + k)$ if $r_i < k$ else $0$, where $r_i$ is a ranking of $b_i$, a value between $0$ and $|\mathcal{B}-1|$.

\subsection{Details of reranker training}\label{app:training-details}

All implementations are based on Tensorflow~\citep{abadi2016tensorflow} and Mesh Tensorflow~\citep{shazeer2018mesh}.
All experiments are done in Google Cloud TPU.
We use batch size that is the maximum that fits one instance of TPU v3-32 (for \webqsp\ and \ambigqa) or TPU v3-8 (\trec).
We use the same batch size for \base; for \citet{nogueira-etal-2020-document}, we use the batch size of 1024.
We use the encoder length of $360$ and the decoder length of $k$ (\sys) or $1$ (all others).
We use $k=\{5, 10\}$ for all experiments. We train \sys\ with $\gamma=\{0, 0.5, 1.0, 1.5\}$ and choose the one with the best accuracy on the \dev\ data.
We use a flat learning rate of $1 \times 10^{-3}$ with warm-up for the first 500 steps.
Full hyperparameters are reported in Table~\ref{tab:hyperparam}. 
\begin{table}[t]
    \centering \small
    \begin{tabular}{l l P{0.5cm} P{1cm} P{0.5cm} P{0.5cm}}
        \toprule
            & $k$ & $B$ & \# train steps & $\gamma$ & $\beta$ \\
        \midrule
            \multirow{2}{*}{\webqsp} & 5 & 256 & \multirow{2}{*}{10k} & 1.5 & 3.0 \\
            & 10 & 224 & & 1.5 & 1.5 \\
        \midrule
            \multirow{2}{*}{\ambigqa} & 5 & 256 & \multirow{2}{*}{6k} & 1.0 & 2.5 \\
            & 10 & 224 & & 1.0 & 2.0 \\
        \midrule
            \multirow{2}{*}{\trec} & 5 & 64 & \multirow{2}{*}{3k} & 1.5 & 1.5 \\
            & 10 & 56 & & 1.5 & 2.0 \\
        \bottomrule
    \end{tabular}
    \caption{Full hyperparamters for training \sys.}\label{tab:hyperparam}
\end{table}

\begin{algorithm}[t]
\caption{An algorithm to obtain $\Tilde{\mathcal{O}}$ from the answer set and $\mathcal{B}$.}\label{alg:preproc}
\begin{algorithmic}[1] \small
\Procedure{\textsc{Preproc}}{$k, \{a_1...a_n\}, \mathcal{B}$}
\State $\Tilde{\mathcal{O}} \gets \empty$ {\protect\color{gitgreen}  {\textit{// a set of positive passages}}} 
\State $\mathcal{A}_\text{left} \gets \{a_1...a_n\}$
\For{$b$ in $\mathcal{B}$}
    \If {$b$ covers any of $\mathcal{A}_\text{left}$}
        \State $\Tilde{\mathcal{O}} \gets \Tilde{\mathcal{O}}$.add($b$)
        \State $\mathcal{A}_\text{left} \gets \mathcal{A}_\text{left} - $ answers in $b$
    \EndIf
    \If {$|\Tilde{\mathcal{O}}|==k$}
        \State \textbf{break}
    \EndIf
\EndFor
\State \Return $O\mathrm{.toSet()}$
\EndProcedure
\end{algorithmic}
\end{algorithm}

\begin{table*}[t]
    \centering \small
    \begin{tabular}{l p{3cm} P{1.3cm} P{1.3cm} P{1.3cm} P{1.3cm} P{1.3cm}}
        \toprule
            \multirow{2}{*}{$k$} & \multirow{2}{*}{Models} &  \multicolumn{2}{c}{\webqsp} &
            \multicolumn{1}{c}{\ambigqa} &
            \multicolumn{2}{c}{\trec} \\
            \cmidrule(lr){3-4} \cmidrule(lr){5-5} \cmidrule(lr){6-7}
            && Dev & Test & Dev & Dev & Test \\
        \midrule
            %\bolden{$k=5$} \\
            \multirow{2}{*}{$5$} & \base & \webqBaseNDCG & \webqBaseTNDCG & \ambigBaseNDCG & \trecBaseNDCG & \trecBaseTNDCG\\
            & \sys & \webqOursNDCG & \webqOursTNDCG & \ambigOursNDCG & \trecOursNDCG & \trecOursTNDCG \\
        \midrule
            %\bolden{$k=10$} \\
            \multirow{2}{*}{$10$} & \base & \webqBaseTenNDCG & \webqBaseTenTNDCG & \ambigBaseTenNDCG & \trecBaseTenNDCG & \trecBaseTenTNDCG\\
            & \sys & \webqOursTenNDCG & \webqOursTenTNDCG & \ambigOursTenNDCG & \trecOursTenNDCG & \trecOursTenTNDCG \\
        \bottomrule
    \end{tabular}
    \caption{
    Results on passage retrieval in $\alpha$-NDCG.
    }\label{tab:results-ndcg}
\end{table*}

\begin{figure*}[ht]
\centering
\resizebox{1.7\columnwidth}{!}{\includegraphics[width=1.7\textwidth]{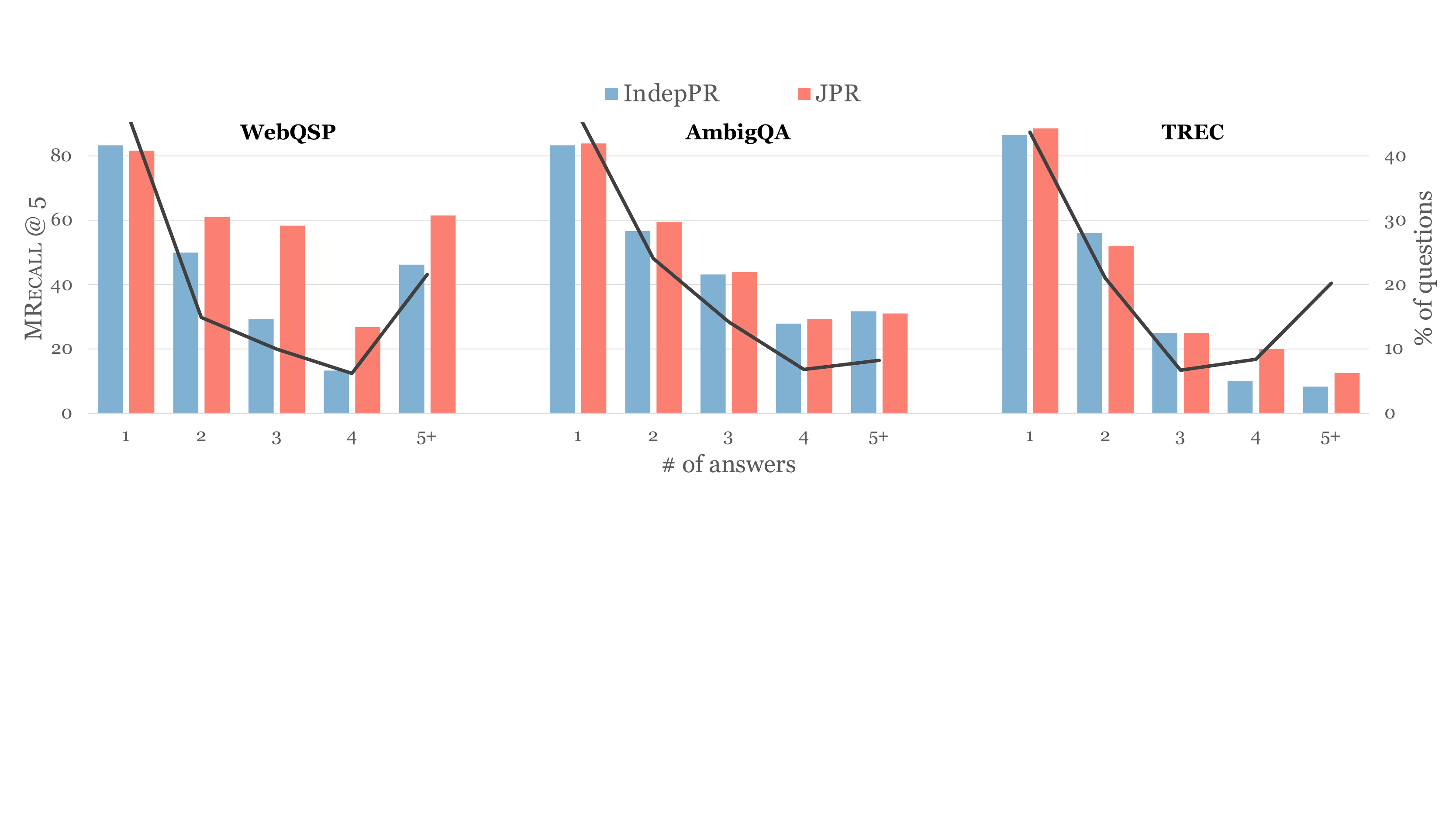}}
\caption{\base\ vs. \sys\ on the \dev\ data of three datasets. %\webqsp\ (left), \ambigqa\ (middle) and \trec\ (right).
\textsc{MRecall} @ $5$ is reported. Lines indicate \% of questions in the data.
\textbf{\sys\ benefits more on questions with 2+ distinct answers.}
}
\label{fig:perf_per_n_answers}\end{figure*}

For training \base\ and \sys, instead of using all of $|\mathcal{B}|$ passages, we use $|\mathcal{B}|/4$ passages by sampling $k$ positive passages and $|\mathcal{B}|/4-k$ negative passages. We find that this trick allows larger batch size when using the same hardware, ultimately leading to substantial performance gains.
We also find that assigning indexes of the passages based on a prior, e.g., ranking from dense retrieval, leads to significant bias, e.g., in 50\% of the cases, the top-1 passage from dense retrieval contains a correct answer. We therefore randomly assign the indexes, and find this gives significantly better performance.

Algorithm~\ref{alg:preproc} describes how a set of positive passages $\Tilde{\mathcal{O}}$ used in Section~\ref{subsection:training} is computed during preprocessing.

\subsection{Details of answer generation training}\label{app:qa-training-details}

We train the models using a batch size of 32. We use a decoder length of $20$ and $40$ for \nq\ and multi-answer datasets, respectively.
We decode answers only when they appear in the retrieved passages, as we want the generated answers to be grounded by Wikipedia passages.
Answers in the output sequence follow the order they appear in the passages, except on \webqsp, where shuffling the order of the answers improves the accuracy.
All other training details are the same as details of reranker training.

\section{Additional Results}\label{app:additional-results}

We additionally report retrieval performance in \bolden{$\alpha$-NDCG @ $k$}, one of the metrics for diverse retrieval in IR~\citep{clarke2008novelty,sakai2019diversity}. It is a variant of NDCG~\citep{jarvelin2002cumulated}, % which considers a cumulative gain for every passage using $k$ as depth,
but penalizes retrieval of the same answer.
We refer to \citet{clarke2008novelty} for a complete definition.
We use $\alpha=0.9$.

Results are reported in Table~\ref{tab:results-ndcg}.
\sys\ consistently outperforms \base\ across all datasets, although the gains are less significant than the gains in \textsc{MRecall}.
We note that we report $\alpha$-NDCG following IR literatures, but we think of \textsc{MRecall} as a priority, because $\alpha$-NDCG does not use an explicit notion of {\em completeness} of retrieval of all answers.
It is also a less strict measure than recall because it gives partial credits to retrieving a subset of the answers.

\paragraph{Gains with respect to the number of answers}
Figure~\ref{fig:perf_per_n_answers} shows gains over \base\ on three datasets with respect to the number of answers. Overall, gains are larger when the number of answers is larger, especially for \webqsp\ and \trec. For \ambigqa, the largest gains are when the number of answers is 2,
which is responsible for over half of multi-answer questions.
%likely because it is responsible for over half of multi-answer questions.

\end{document}